\documentclass[aps,pre,preprint,superscriptaddress,longbibliography,floatfix]{revtex4-2}
\usepackage{amsmath,amssymb,bm,mathtools}
\usepackage{graphicx}
\usepackage{booktabs}
\usepackage{tikz}
\usetikzlibrary{arrows.meta,positioning,fit,backgrounds,calc,shapes.geometric}
\usepackage{placeins}
\usepackage{enumitem}
\usepackage[dvipsnames]{xcolor}
\usepackage[hidelinks]{hyperref}
\usepackage{url}
\setcitestyle{numbers,square,sort&compress}
\providecommand{\R}{\mathbb{R}}
\providecommand{\E}{\mathbb{E}}

\newcommand{\GMM}{\mathrm{GMM}}
\newcommand{\code}[1]{\texttt{#1}}

\providecommand{\dd}{\,\mathrm{d}}
\providecommand{\tr}{\mathrm{tr}}
\providecommand{\n}{\nabla}
\providecommand{\Del}{\Delta}
\newtheorem{theorem}{Theorem}

\newenvironment{revision}{\color{black}}{\color{black}}

\usepackage{comment}

\begin{document}
\title{Analytic Bridge Diffusions for Controlled Path Generation}
\author{Michael (Misha) Chertkov}
\email{chertkov@arizona.edu}
\affiliation{Graduate Interdisciplinary Program in Applied Mathematics \& Department of Mathematics, University of Arizona, Tucson, Arizona 85721, USA}
\date{\today}
\begin{abstract}
\begin{revision}
Most modern bridge-diffusion methods achieve finite-time transport by specifying an interpolation, Schrödinger-bridge, or stochastic-control objective and then learning the associated score or drift field with a neural network. In contrast, we identify a restricted but sufficiently broad analytically solvable class in which, for a deterministic source and a Gaussian-mixture target, the score and all intermediate marginals are explicit and protocol objectives of the type used in this paper can be differentiated without inner stochastic simulation loops. We recast the classical linear--quadratic--Gaussian stochastic-control structure as a transport problem of the Path Integral Diffusion  type. Linear dynamics, Gaussian noise, and quadratic running costs reduce the bridge calculation to a matrix Riccati cascade, while the terminal state cost is replaced by a prescribed Gaussian-Mixture terminal probability density.

Linear Quadratic -- Gaussian Mixture -- Path Integral Diffusion (LQ-GM-PID) thereby turns bridge diffusion from terminal target matching alone into an analytically controlled laboratory for path shaping. We demonstrate this on a 2D corridor task, a 2D multi-entrance task, and a high-dimensional study reaching $d=32$ and $M=16$ terminal modes in separate scaling sweeps. We position LQ-GM-PID as an analytically solvable reference model in which score approximations, path-shaping objectives, and protocol-learning procedures can be tested against explicit quantities.
\end{revision}

\end{abstract}

\maketitle

\section{Introduction}
\label{sec:intro}

Bridge diffusions --- finite-time, non-autonomous transports between a tractable reference and a target distribution --- have become a unifying abstraction for generative modeling. Score-based diffusions \citep{ho_denoising_2020, song_score-based_2021}, denoising bridges \citep{peluchetti_non-denoising_2021, peluchetti_diffusion_2023}, Schr\"odinger bridges \citep{schrodinger_sur_1932, de_bortoli_diffusion_2021, chen_likelihood_2023, shi_diffusion_2023}, stochastic interpolants \citep{albergo_building_2023, albergo_stochastic_2025}, and flow matching \citep{lipman_flow_2023} can all be cast as different parameterizations of the same underlying problem: choose a drift field $u_t(x)$ on $[0,1]\times\R^d$ such that the controlled stochastic differential equation (SDE) $dX_t = u_t(X_t)\,dt+dW_t$ produces a prescribed terminal distribution and, secondarily, traces a useful intermediate path. The dominant practical recipe parameterizes $u_t$ by a neural network and trains it by score matching, denoising score matching, or maximum-likelihood. The expressive power is enormous; the analytic transparency is essentially zero.

\begin{revision}
This paper takes the opposite stance. We ask: \emph{within the bridge-diffusion family, is there a sufficiently expressive sub-family in which the optimal drift and intermediate marginals are explicit and useful path-level protocol objectives can be optimized without learning the score?} Our answer is \textbf{Linear Quadratic -- Gaussian Mixture -- Path Integral Diffusion} (LQ-GM-PID), a Path Integral Diffusion (PID) model obtained by promoting the classical linear--quadratic--Gaussian (LQG) stochastic-control structure to a finite-time transport problem. Classical LQG control combines linear dynamics, Gaussian noise, and quadratic costs; its solvability is encoded by Riccati equations and linear optimal feedback laws \citep{kalman_contributions_1960,kalman_new_1961,wonham_matrix_1968,kwakernaak_linear_1972,anderson_optimal_1990}. In LQ-GM-PID we retain the linear--quadratic stochastic-control backbone but replace terminal state regulation by a prescribed Gaussian-Mixture (GM) terminal density. The exact closed-form theorem is stated for a deterministic source. A non-degenerate source is treated in the experiments by a finite empirical approximation: each sampled initial point defines a deterministic-start conditional bridge. The protocol $\Gamma_t=(\beta_t,\nu_t,\sigma_t,\kappa_t)$ remains low-dimensional and piecewise constant in time: matrix $\beta_t$ controls anisotropic path stiffness, vector $\nu_t$ specifies the moving guide, matrix $\sigma_t$ defines a linear state-dependent base drift, and scalar $\kappa_t$ controls stochastic spread.

The technical core is therefore not merely the replacement of a scalar stiffness by a matrix. It is the combination of (i) a self-contained matrix Riccati construction in arbitrary dimension, (ii) explicit optimal score and intermediate Gaussian-mixture  marginals for deterministic-source/Gaussian-mixture-target bridges, and (iii) density-level differentiation of path objectives through this analytic backbone. The empirical studies then exercise these objects in progressively richer geometries: corridor tracking, sample-conditioned multi-entrance transport, and high-dimensional coarse-to-fine branching. None uses a neural network in the optimization loop.
\end{revision}

\begin{revision}
\paragraph{What ``no neural networks'' buys us.} Within the deterministic-source/Gaussian-mixture-target class of Theorem~\ref{thm:closedform}, closed-form structure delivers four useful properties: (a)~exact terminal matching in continuous time; (b)~explicit intermediate marginals, which make the Gaussian-kernel path objectives used in E1/E2 experiments analytically evaluable; (c)~differentiability through a finite Riccati cascade rather than through a learned score or an unrolled SDE simulation (the deterministic-source cascade scales as $O(Kd^3)$, with E2 adding a finite average over frozen source atoms); and (d)~interpretability of the PWC protocol $\Gamma_t=(\beta_t,\nu_t,\sigma_t,\kappa_t)$. For E2, statements (a)--(c) apply conditionally to each deterministic source atom and are then averaged over the empirical source. We make no claim that LQ-GM-PID supersedes neural diffusion models on perceptual generative tasks; the aim is an analytic baseline and construction on which extensions can be tested.
\end{revision}

\begin{figure}[htbp]
\centering
\resizebox{0.98\linewidth}{!}{%
\begin{tikzpicture}[
  font=\small,
  every node/.style={transform shape},
  >={Stealth[length=2mm]},
  block/.style={draw, rounded corners=2pt, align=center,
    minimum height=1.0cm, minimum width=2.0cm, fill=white},
  inputbox/.style={draw, fill=blue!8, rounded corners=1pt, align=center,
    minimum height=0.8cm, minimum width=1.7cm},
  outbox/.style={draw, fill=orange!10, rounded corners=1pt, align=center,
    minimum height=0.8cm, minimum width=2.0cm},
  corebox/.style={draw, very thick, fill=black!4, rounded corners=3pt,
    align=center, minimum height=2.4cm, minimum width=4.5cm},
  arr/.style={->, thick},
  panel/.style={draw, dashed, rounded corners=4pt, inner sep=8pt}
]

\node[inputbox] (src) at (0,1.6) {Source\\$\delta(x-x_0)$ or finite empirical $\{z^{(n)}\}$};
\node[inputbox] (proto) at (0,0.3) {Protocol $\Gamma_t$\\$(\beta_t,\nu_t,\sigma_t)$};
\node[inputbox] (tgt) at (0,-1.0) {Target $p^{\mathrm{tar}}(x)$\\$\GMM_{\mathrm{tar}}$};

\node[draw, dashed, rounded corners=4pt, inner sep=6pt,
      fit=(src)(proto)(tgt), label=above:\textbf{A. Inputs}] (panelA) {};

\node[corebox] (core) at (6.6,0.3)
  {\textbf{LQ-GM-PID}\\[2pt]
   Backward Riccati  $\dot A^-, \dot B^-, \dot C^-$\\
   Forward Riccati  $\dot A^+, \dot B^+$\\[2pt]
   $u^*_t(x) = \nabla_x \log\psi_t(x)$\\
   $p_t(x) = \sum_k \bar w_k\, \mathcal{N}(\mu_k, \Sigma_k)$};

\node[draw, dashed, rounded corners=4pt, inner sep=6pt,
      fit=(core), label=above:\textbf{B. Analytic backbone}] (panelB) {};

\draw[arr] (src.east) -- ++(0.4,0) |- (core.west |- src);
\draw[arr] (proto.east) -- (core.west |- proto);
\draw[arr] (tgt.east) -- ++(0.4,0) |- (core.west |- tgt);

\node[outbox] (e1) at (12.4,1.6)  {E1 corridor\\(2D, learned $\nu_t$)};
\node[outbox] (e2) at (12.4,0.3)  {E2 finite-source\\multi-entrance transport};
\node[outbox] (h1) at (12.4,-1.0) {H1 scaling\\$d\leq32, M\leq16$};

\node[draw, dashed, rounded corners=4pt, inner sep=6pt,
      fit=(e1)(e2)(h1), label=above:\textbf{C. Experiments}] (panelC) {};

\draw[arr] (core.east |- e1) -- (e1.west);
\draw[arr] (core.east |- e2) -- (e2.west);
\draw[arr] (core.east |- h1) -- (h1.west);

\end{tikzpicture}%
}
\caption{\textbf{LQ-GM-PID at a glance.}
\begin{revision}\textbf{(A)~Inputs.} A deterministic source (or a finite empirical set of source points used conditionally in E2), a Gaussian-mixture target, and a piecewise-constant protocol $\Gamma_t = (\beta_t, \nu_t, \sigma_t)$. \textbf{(B)~Closed-form analytic backbone.} For each deterministic start, the forward and backward Kolmogorov--Fokker--Planck operators reduce to coupled matrix Riccati systems. The optimal score and instantaneous marginal are explicit; path objectives such as the corridor-overlap objective used below are differentiated through the same cascade. \textbf{(C)~Empirical demonstrations.} E1 learns a 2D corridor protocol, E2 averages deterministic-start conditional bridges over a fixed empirical sample from a bimodal source, and H1 studies structured branching up to $d{=}32$ and $M{=}16$.\end{revision}}
\label{fig:schematic}
\end{figure}

\section{Related work}
\label{sec:related}

\paragraph{Score-based diffusion, bridge-diffusion samplers, and optimal transport.} 

Modern diffusion-based generative modeling \citep{sohl-dickstein_deep_2015,ho_denoising_2020, song_score-based_2021} parameterizes the score field by a neural network and trains by score matching. The bridge-diffusion viewpoint \citep{peluchetti_non-denoising_2021, peluchetti_diffusion_2023} reframes sampling as a finite-time, non-autonomous transport between a tractable reference and a target. Neural Schr\"odinger-bridge approaches \citep{de_bortoli_diffusion_2021, chen_likelihood_2023, shi_diffusion_2023,liu_generalized_2023, vargas_solving_2021} target the entropic-optimal-transport problem with iterative proportional fitting or forward--backward SDE schemes; stochastic interpolants \citep{albergo_building_2023, albergo_stochastic_2025} and flow matching \citep{lipman_flow_2023} provide simulation-free training objectives for deterministic bridges. 

The pre-history of the underlying transport problem reaches back to \citet{monge_memoire_1781} and \citet{kantorovich_translocation_1942}. Schr\"odinger's bridge problem \citep{schrodinger_sur_1932} introduced the entropic, stochastic counterpart of optimal transport: among path measures consistent with prescribed endpoint marginals, choose the one closest in relative entropy to a reference diffusion. This viewpoint connects optimal transport to stochastic control, reciprocal processes, and entropy-regularized transport; see, for example, the modern synthesis in \citet{chen_stochastic_2021}. In the machine-learning literature, this control-theoretic interpretation appears already in finite-time latent diffusion models for sampling and inference, where the drift of a diffusion is chosen or learned so that its terminal law matches a target distribution \citep{tzen_theoretical_2019}. Neural Schr\"odinger bridges, stochastic interpolants, flow matching, and related bridge-diffusion samplers can be read as modern incarnations of this same finite-time transport idea.

Most of the modern methods named above share a common template: choose a bridge, interpolation, or stochastic-control objective; parameterize the score or drift field by a flexible function class, typically a neural network such as a U-Net, feed-forward neural network, or transformer; and optimize the resulting training objective from samples. \begin{revision}LQ-GM-PID occupies a complementary position. For a deterministic source and Gaussian-mixture target, we restrict the protocol to a linear--quadratic subclass for which the optimal drift and intermediate marginals are explicit; the Gaussian-kernel density-level objectives used in this paper are then analytic functions of those marginals.\end{revision}

\paragraph{Path Integral Control, Path Integral Diffusion, and the Harmonic line.} The closed-form structure of LQ-GM-PID descends from a deep integrable substructure of stochastic optimal control known as Path Integral Control (PIC) -- introduced in \cite{kappen_path_2005} but with a longer pre-history in the control community, e.g. \cite{fleming_exit_1977,mitter_non-linear_1981,fleming_optimal_1982}, generalization known under the name of linearly solvable Markov decision process \cite{todorov_linearly-solvable_2007,e_todorov_general_2008,todorov_efficient_2009,dvijotham_unifying_2012,dvijotham_linearly_2013} and more-recent developments and follow ups \cite{theodorou_relative_2012,kappen_optimal_2012,chernyak_stochastic_2014,kappen_adaptive_2016}. \citet{kappen_linear_2005} observed that for stochastic optimal-control (SOC) problems with linear, additive control entering through the same channels as the noise and with the cost being an integral over time with integrand split into quadratic in control term and an arbitrary state dependent term, a Hopf--Cole transformation collapses the nonlinear Hamilton--Jacobi--Bellman equation to a linear backward Kolmogorov--Fokker--Planck equation, so the value function is expressible as a Feynman--Kac path integral against the uncontrolled process. \citet{kappen_adaptive_2016} developed adaptive importance sampling based on PIC. The PIC viewpoint was deepened and given a calculus-of-variations formulation over densities and currents in \citet{chernyak_stochastic_2014}. 

Path Integral \emph{Diffusion} (PID) --- the use of this analytic backbone for endpoint-constrained optimal transport and respective sampling rather than for control --- and its Harmonic specialization (H-PID, scalar time-independent $\beta$, sampled via importance sampling) were introduced in \citet{behjoo_harmonic_2025}. Related harmonic/Schr\"odinger-bridge structure was also explored independently by \citet{teter_schrodinger_2024}, providing a complementary perspective on analytically tractable bridge constructions. \begin{revision}\citet{chertkov_adaptive_2026} (Adaptive Path Integral Diffusion -- AdaPID) extended H-PID to time-dependent scalar $\beta$ and considered non-degenerate initial ensembles; the present construction distinguishes explicitly between deterministic-start closed forms and sample-conditioned treatment of such ensembles. \citet{chertkov_generative_2025} added a moving centerline/guide $\nu_t$ in two dimensions.\end{revision}

\begin{revision}The present paper completes this line by generalizing to matrix-valued $\beta_t$, partial $\sigma_t$ for state-dependent linear drift, and arbitrary ambient dimension, while making the deterministic-source analytic backbone explicit and self-contained --- including a closed-form expression for the instantaneous marginal $p_t(x)$ as a Gaussian mixture for every $t\in(0,1)$ when the target is a Gaussian mixture.\end{revision} Recent reward-fine-tuning work casts diffusion as SOC and uses adjoint methods for gradient flow \citep{domingo-enrich_adjoint_2024}; LQ-GM-PID's analytic Riccati cascade provides a closed-form alternative to the adjoint within its restricted class.

\paragraph{Sampling paths, not just endpoints.} The bridge-diffusion enterprise often cares about the trajectory ensemble itself, not only its terminal law. The canonical motivation is sampling \emph{transition paths} between metastable states of a molecular system --- folded vs.\ unfolded protein, reactants vs.\ products of a chemical reaction --- where standard molecular dynamics is dominated by waiting time at the metastable basins and almost never crosses the barrier within a feasible simulation budget. \citet{holdijk_stochastic_2023} related molecular transition path sampling to the Schr\"odinger bridge problem and to stochastic optimal control, learning the controller with a neural-network policy. Subsequent neural approaches include the variational Doob's $h$-transform of \citet{du_doobs_2024}, the diffusion-based path generation of \citet{triplett_diffusion_2023}, and the action-minimization framework of \citet{raja_action-minimization_2025}. Path sampling motivates the present work's emphasis on the entire trajectory $\{X_t\}_{t\in[0,1]}$: the LQ-GM-PID protocol $\Gamma_t = (\beta_t, \nu_t, \sigma_t)$ is a path-shaping object, and the diagnostics in \S\ref{sec:experiments} measure path quality, not only terminal fidelity.

\paragraph{Detailed balance versus control of the transient.} Two structurally distinct routes to sampling from a target density $p^{\mathrm{tar}}\propto e^{-E}$ are visible in the literature. \begin{revision}The Langevin / Markov-chain Monte Carlo (MCMC) family designs a stationary Markov process whose invariant measure is $p^{\mathrm{tar}}$ and relies on detailed balance for asymptotic exactness; structured decompositions of the energy $E$ can make otherwise intractable joint distributions easier to sample.\end{revision} The price is that exactness is achieved only as $t\to\infty$, with rate governed by the spectral gap, which can be exponentially small in the presence of metastability. The bridge-diffusion family takes the opposite stance: it abandons detailed balance, gives up stationarity, and instead designs a non-autonomous, finite-horizon transport whose marginal at $t=1$ matches $p^{\mathrm{tar}}$ exactly --- in unit time. The present paper sits in this second stance and makes it more explicit: we exploit \emph{control of the dynamic transient} to deliver target sampling without ever appealing to detailed balance. Several recent works can be read in the same spirit: \citet{chen_sequential_2024} couple sequential Monte Carlo with controlled Langevin diffusions; \citet{guo_gaussian_2023} replace the Gaussian transition kernel of standard diffusion solvers by a Gaussian mixture kernel, motivated by the observation that the Gaussian assumption is violated for mixture data; \citet{dyachenko_variational_2026} note that closed-form Gaussian-mixture parameterizations of entropic OT are an alternative to neural parameterizations; \citet{du_doobs_2024} parameterize Doob's $h$-transform variationally with simulation-free training. LQ-GM-PID is the most aggressive entry in this thread: the analytic Gaussian-mixture machinery is not a kernel approximation or a parameterization choice but the entire backbone, with the optimal drift itself --- not just the transition kernel --- available in closed form on the LQ-GM subclass.

\paragraph{Closed-form Gaussian Mixture models as baselines.} \citet{guo_gaussian_2023}, \citet{dyachenko_variational_2026}, and \citet{du_doobs_2024} use Gaussian-mixture structure to obtain partial analytic tractability inside otherwise neural pipelines. \begin{revision}LQ-GM-PID pushes this analytic program to the full deterministic-source/Gaussian-mixture-target bridge: the optimal control and every intermediate marginal are explicit, and the Gaussian-kernel protocol objective used in E1 can be differentiated through that analytic backbone. E2 retains the same conditional formulas but averages them over a finite empirical source.\end{revision} We position LQ-GM-PID accordingly: a sub-family small enough to admit full analytic treatment but large enough to exhibit the qualitative phenomena that motivate the field. The mean-field extension to interacting agents \citep{chertkov_mean-field_2026} is discussed in \S\ref{sec:discussion}.

\section{Our Contributions}
\label{sec:contributions}

\begin{revision}
The paper makes four contributions.
\begin{enumerate}[leftmargin=1.5em]
\item \textbf{An explicit LQ-GM bridge backbone (\S\ref{sec:lqgmpid}, App.~\ref{app:derivations}).} For a deterministic source and Gaussian-mixture target, linear drift and quadratic path guidance reduce both Green functions to a matrix Riccati cascade. This yields explicit component-wise formulas for the optimal score and for the full intermediate marginal at every $t\in(0,T)$ in arbitrary dimension.
\item \textbf{Density-level path-objective learning (\S\ref{sec:objective}, App.~\ref{app:caseB}).} The explicit marginal makes Gaussian/quadratic path-overlap objectives differentiable through the Riccati cascade without an inner trajectory-simulation loop. We illustrate it optimizing the $20$ piecewise-constant parameters $(\rho_k,c_k)_{k=1}^{10}$ of a 2D anisotropic corridor protocol.
\item \textbf{Sample-conditioned multi-entrance transport (\S\ref{sec:E2}, App.~\ref{app:caseC}).} For a non-degenerate source we draw a fixed empirical sample $\{z^{(n)}\}_{n=1}^B$ and apply the deterministic-source formulas conditionally to each $z^{(n)}$. The resulting finite mixture remains analytically evaluable and supports the same corridor objective. We explicitly distinguish this empirical-source construction from the continuous two-marginal bridge.
\item \textbf{High-dimensional structured transport (\S\ref{sec:H1}, App.~\ref{app:H1}).} At $d\le32$ and $M\le16$, hand-crafted matrix-valued $\beta_t$ protocols produce anisotropic confinement, delayed branch release, and a trunk--branch--local variance hierarchy whose analytic marginal agrees with sampled trajectories.
\end{enumerate}
\end{revision}

\begin{revision}
\paragraph{Scope and relation to the earlier work.} We do not claim novelty for linearly solvable stochastic optimal control itself, nor that this restricted class competes with neural generative models on perceptual benchmarks. Relative to the earlier scalar guided harmonic construction, the advance here is the explicit matrix-valued, arbitrary-dimensional bridge calculation together with the exact intermediate marginal and its use for density-level path-protocol learning. The closed-form source theorem is deterministic-start; a genuine non-degenerate two-marginal source requires an additional endpoint scaling and lies outside the analytic theorem proved here. The $\sigma_t$-dependent Riccati equations are retained in the analytic derivation, while systematic protocol learning with nonzero $\sigma_t$ is left for future work.
\end{revision}

\section{The LQ-GM-PID class}
\label{sec:lqgmpid}

This section makes the analytic class precise and states the closed-form result on which the rest of the paper rests. Derivations are deferred to App.~\ref{app:derivations}.

\paragraph{The controlled SDE.} Continuing the line of work on Harmonic Path Integral Diffusion (H-PID) \cite{behjoo_harmonic_2025,chertkov_adaptive_2026,chertkov_generative_2025} we work on the time interval $[0,T]$ with $T=1$ in all experiments and study the (controlled) stochastic process described via the It\^o SDE
\begin{align} 
  \dd X_t &= \bigl(f_t(X_t)+u_t(X_t)\bigr)\dd t + \sqrt{\kappa_t}\,\dd W_t,\quad X_0 \sim p^{(\mathrm{in})}, \quad X_t\in\R^d,
  \label{eq:sde_lqgmpid}
\end{align}
where $\kappa_t>0$ is a scalar (time-dependent) diffusion coefficient, and the cost-to-go (value function) is defined via the sliding time optimization
\begin{align}
  J_t(x)
  &= \inf_{u_{t\to {T}}}
     \E\Bigl[
        \int_t^{T} \Bigl(\frac{1}{2\kappa_{t'}} \|u_{t'}(X_{t'})\|^2 + V_{t'}(X_{t'})\Bigr)\dd t'
        + \phi(X_{T})
        \,\Big|\, X_t=x
     \Bigr],
  \label{eq:cost_to_go}
\end{align}
where (as discussed in details in App.~\ref{appA:general_pid}) the target cost $\phi$ can also be substituted by the conjugated terminal/target law 
\begin{equation}\label{eq:tar}
X_T\sim p^{(\mathrm{tar})}.
\end{equation}
LQ-GM-PID is the subclass defined by three structural restrictions on the triple $(f_t, V_t, p^{(\mathrm{tar})})$:
\begin{equation}
  f_t(x) \;=\; \sigma_t\, x, \qquad V_t(x) \;=\; \tfrac{1}{2}(x-\nu_t)^\top \beta_t (x-\nu_t),
  \qquad
  p^{(\mathrm{tar})}(y) \;=\; \sum_{k=1}^{K}\pi_k\,\mathcal N(y;\,m_k,\Sigma_k),
  \label{eq:lqgm_specification}
\end{equation}
\begin{revision}where $\sigma_t \in \R^{d\times d}$ is time dependent, $\beta_t \in \R^{d\times d}$ is symmetric positive definite, and $\nu_t\in \R^d$ is a moving centerline. The closed-form theorem below assumes a deterministic source $p^{(\mathrm{in})}=\delta(\cdot-x_0)$. Section~\ref{sec:E2} later uses a finite empirical source sampled from a Gaussian mixture and averages deterministic-start conditional quantities; it does not assert an analytic solution of the continuous Gaussian-mixture-to-Gaussian-mixture bridge. Throughout we use piecewise-constant (PWC) protocols $\Gamma_t=(\beta_t,\nu_t,\sigma_t,\kappa_t)$ on $0=t_0<\cdots<t_K=T$.\end{revision}

\paragraph{Why these four pieces.} The LQ guide potential $V_t$ is the smallest non-trivial choice that preserves the linearity of the backward Hamilton--Jacobi--Bellman equation under the Hopf--Cole transform \citep{behjoo_harmonic_2025}. The matrix $\beta_t$ controls \emph{anisotropic corridor stiffness} --- it can be loose along directions in which the cloud is allowed to spread and tight along directions in which the cloud must remain concentrated. The centerline $\nu_t$ specifies the moving target along which path adherence is measured; the linear drift $\sigma_t x$ introduces a state-dependent contraction or expansion direction; and $\kappa_t$ schedules the diffusion strength. Restrictive though Eq.~(\ref{eq:lqgm_specification}) appears, the Gaussian-mixture target absorbs arbitrary multi-modal terminal structure, the matrix-valued $\beta_t$ absorbs arbitrary anisotropic corridor geometry, and the moving $\nu_t$ absorbs arbitrary path-shaping demands. The specialization of GuidedPID to scalar $\beta$ and zero $\sigma$ \citep{chertkov_generative_2025} is the strict subclass with $\beta_t= \beta_t I$ and $\sigma_t = 0$; AdaPID \citep{chertkov_adaptive_2026} relaxes only the time-dependence of scalar $\beta$.

\begin{revision}
\paragraph{Construction in five steps.}
The analytic logic can be read as follows. (1) The quadratic control cost gives $u_t^*(x)=\kappa_t\nabla\log\psi_t(x)$ after the Hopf--Cole transform $\psi_t=e^{-J_t}$. (2) The transformed Hamilton--Jacobi--Bellman equation is linear, with backward Green function $G_t^{(-)}(x\mid y)$; the adjoint equation has forward Green function $G_t^{(+)}(x\mid x_0)$. (3) For a deterministic source, terminal matching is imposed by the explicit reweighting
\begin{equation}
\psi_t(x)\propto \int_{\mathbb R^d} p^{(\mathrm{tar})}(y)\,
\frac{G_t^{(-)}(x\mid y)}{G_T^{(+)}(y\mid x_0)}\,\dd y,
\qquad
p_t^*(x)\propto G_t^{(+)}(x\mid x_0)\psi_t(x),
\label{eq:pedagogical_ratio}
\end{equation}
which already explains the terminal look-up rule below. (4) Under linear drift and quadratic $V_t$, both Green functions are Gaussian and their coefficients obey matrix Riccati equations; a PWC protocol therefore gives a finite cascade of matrix exponentials. (5) If $p^{(\mathrm{tar})}$ is a Gaussian mixture, the integral in Eq.~(\ref{eq:pedagogical_ratio}) is a finite sum of Gaussian integrals, producing Theorem~\ref{thm:closedform}.

\paragraph{One-dimensional sanity check.}
Take $d=1$, $\sigma=\beta=0$, $\kappa=1$, $x_0=0$, and $p^{(\mathrm{tar})}=\mathcal N(m,1)$. The passive terminal law is $G_T^{(+)}(y\mid0)=\mathcal N(0,1)$, hence the terminal ratio is proportional to $\exp(my-m^2/2)$. The backward heat semigroup gives $\psi_t(x)\propto\exp(mx-m^2t/2)$, so $u_t^*(x)=\partial_x\log\psi_t=m$ and $p_t^*=\mathcal N(mt,t)$. Thus $p_1^*=\mathcal N(m,1)$ exactly. The full LQ-GM construction is the matrix/Gaussian-mixture generalization of this elementary calculation.
\end{revision}

\paragraph{Closed-form score.} The central technical claim of the paper is that, on the LQ-GM class Eq.~(\ref{eq:lqgm_specification}), the optimal drift, the instantaneous marginal density, and their gradients with respect to the protocol $\Gamma$ are all available in closed form on a single analytic backbone --- the matrix Riccati cascade in App.~\ref{app:derivations}.

\begin{theorem}[LQ-GM-PID closed-form score and marginal]\label{thm:closedform} Let $X_t \in \R^d$ evolve under Eq.~\ref{eq:sde_lqgmpid} with the LQ specialization Eq.~(\ref{eq:lqgm_specification}) and a deterministic source $p^{(\mathrm{in})} = \delta(\cdot - x_0)$. For any PWC protocol $\Gamma$, the optimal drift on $t \in (0, T)$ admits the closed form
\begin{equation}
  u^*_t(x)
  \;=\;
  \kappa_t \sum_{k=1}^{K} \rho_{k,t}(x)\,
  \bigl(-\Lambda_{k,t}\, x + \lambda_{k,t}\bigr),
  \label{eq:u_star_main}
\end{equation}
where $\rho_{k,t}(x) \in (0,1)$ are softmax responsibilities Eq.~(\ref{appA:eq:rho_def}) and the per-component matrices $\Lambda_{k,t} \in \R^{d\times d}$ and vectors $\lambda_{k,t} \in \R^d$ are explicit algebraic functions of: (i)~the backward Riccati outputs $(A^{(-)}_t, B^{(-)}_t, C^{(-)}_t, \theta^{(-)}_{x;t}, \theta^{(-)}_{y;t})$ at the query time $t$; (ii)~the terminal forward outputs $(A^{(+)}_T, B^{(+)}_T, \theta^{(+)}_{x;T})$; and (iii)~the target component parameters $(\pi_k, m_k, \Sigma_k)$. Furthermore, the optimal marginal density is itself a Gaussian mixture for every $t \in (0,T)$:
\begin{equation}
  p^*_t(x) \;=\; \sum_{k=1}^{K} \bar\pi_{k,t}\,\mathcal N\bigl(x;\,\mu_{k,t},\,\Pi_{k,t}^{-1}\bigr),
  \label{eq:p_star_main}
\end{equation}
where $\Pi_{k,t}, \mu_{k,t}, \bar\pi_{k,t}$ are explicit functions of the same backward and forward Riccati outputs. Both Eq.~(\ref{eq:u_star_main}) and Eq.~(\ref{eq:p_star_main}) are reproduced from App.~\ref{app:derivations}, where the per-component quantities $\Lambda_{k,t}, \lambda_{k,t}, \Pi_{k,t}, \mu_{k,t}, \bar\pi_{k,t}$ are given in Eqs.(\ref{appA:eq:Lambda_kt}--\ref{appA:eq:lambda_kt}) and~Eqs.~(\ref{appA:eq:Pi_kt}--\ref{appA:eq:pi_bar}) respectively.

\end{theorem}

\paragraph{Terminal look-up law.}
A second closed-form object, used implicitly by the score formula, is the conditional law of the terminal endpoint given an intermediate state. For a deterministic source, define
\begin{equation}
  \ell_t(y\mid x)
  \doteq
  \frac{\displaystyle p^{(\mathrm{tar})}(y)\,G_t^{(-)}(x\mid y) / G_T^{(+)}(y\mid x_0)}
       {\displaystyle \int_{\mathbb R^d} p^{(\mathrm{tar})}(y')\,G_t^{(-)}(x\mid y') / G_T^{(+)}(y'\mid x_0)\,\dd y'} .
  \label{eq:lookup_law_main}
\end{equation}
We call this the terminal \emph{look-up law}: it is the bridge-induced distribution over possible terminal components and terminal positions that are compatible with the observation $X_t=x$. Its expectation
\begin{equation}
  \widehat x_T(t;x) \doteq \int_{\mathbb R^d} y\,\ell_t(y\mid x)\,\dd y
  \label{eq:lookup_expectation_main}
\end{equation}
provides an inference-time ``look-up'' map from the present state to the implied terminal endpoint. In the Gaussian-mixture case, both $\ell_t(y\mid x)$ and $\widehat x_T(t;x)$ are explicit Gaussian-mixture quantities (App.~\ref{app:derivations}, \S\ref{appA:lookup_law}). This object is useful beyond evaluating the control: its temporal transition from diffuse to component-resolved behavior is often sharper than the marginal motion itself, making it a natural candidate order parameter for the dynamic phase-transition phenomena discussed for H-PID and U-turn diffusion \citep{behjoo_harmonic_2025,biroli_dynamical_2024,behjoo_u-turn_2025}. We return to this point in \S\ref{sec:discussion} as a path toward inference-time and phase-transition diagnostics.

\paragraph{The Riccati cascade.} The matrices and vectors $(A^{(\pm)}_t, B^{(\pm)}_t, C^{(\pm)}_t, \theta^{(\pm)}_{x;t}, \theta^{(\pm)}_{y;t})$ that feed Eqs.~(\ref{eq:u_star_main}--\ref{eq:p_star_main}) are the coefficients of a Gaussian ansatz for the backward and forward Green functions of the linearized Hamilton--Jacobi--Bellman equation. Substituting the ansatz into the linearized HJB and matching coefficients gives a matrix Riccati system for $A^{(\pm)}_t$ and a coupled linear system for the remaining coefficients (App.~\ref{app:derivations}, \S\ref{appA:odes}). Under PWC protocols, the Riccati system has an explicit matrix-exponential solution on each interval, and continuity at the breakpoints chains them into a closed-form cascade (App.~\ref{app:derivations}, \S\ref{appA:pwc_closed_form}). The cost of one cascade evaluation --- forward sweep, backward sweep, and the per-component algebra in Eq.~(\ref{eq:u_star_main}) --- is $O(K\cdot d^3)$, dominated by the $K$ block matrix exponentials of size $2d\times 2d$.

\begin{revision}
\paragraph{What is exact, what is approximate.} For a deterministic source, Theorem~\ref{thm:closedform} is exact in continuous time: the terminal law is $p^{(\mathrm{tar})}$, the intermediate marginal is the Gaussian mixture in Eq.~(\ref{eq:p_star_main}), and the score in Eq.~(\ref{eq:u_star_main}) is explicit. Numerical trajectory plots add only the boundary cutoff and time-discretization error. For E2 experiments, there is an additional and conceptually different approximation: a non-degenerate source is replaced by a fixed empirical measure $B^{-1}\sum_n\delta(\cdot-z^{(n)})$, and the deterministic-start construction is applied conditionally to each atom. Conditional on that finite empirical source, the quantities used in the optimization are analytic; as $B\to\infty$ the empirical source converges to the prescribed source law. This construction is not the exact Markov bridge associated with two continuous endpoint marginals.
\end{revision}

\section{Density-level protocol comparison and learning}
\label{sec:objective}

The closed-form marginal Eq.~(\ref{eq:p_star_main}) of Theorem~\ref{thm:closedform} is what makes protocol comparison, and ultimately protocol \emph{learning}, tractable in this paper. Where neural-network bridge diffusions train a parametric drift to approximate the score, we compare and learn over a low-dimensional \emph{protocol} $\Gamma$ whose induced score is exact at every step, and we do so by gradient methods that never simulate the SDE.

\begin{revision}
\paragraph{Density-level objectives.}
Eq.~(\ref{eq:p_star_main}) makes a useful class of path objectives analytic. A schematic decomposition is
\begin{equation}
J(\Gamma)=J_{\mathrm{path}}(\Gamma)+\alpha_{\mathrm{kin}}J_{\mathrm{kin}}(\Gamma)+\lambda_{\mathrm R}\mathcal R(\Gamma),
\label{eq:Jpath}
\end{equation}
where $J_{\mathrm{path}}$ measures adherence to a desired route, $J_{\mathrm{kin}}$ may penalize control effort, and $\mathcal R$ penalizes undesirable variation of the PWC protocol. Equation~(\ref{eq:Jpath}) is a template, not a prescription to optimize all of its coefficients. Moreover, not every possible integrand is reducible to Gaussian second moments: in particular, the squared norm of the mixture score generally retains the soft responsibilities and may require numerical quadrature if it is used.

The E1/E2 experiments use the narrower objective that is actually analytic and is sufficient for the present demonstration:
\begin{equation}
\mathcal L(\Gamma)=10\,\mathcal L_{\mathrm{corr}}(\Gamma)+0.10\,\mathcal L_{\rho}(\Gamma)+0.05\,\mathcal L_{\beta}(\Gamma),
\label{eq:actual_protocol_loss}
\end{equation}
with all three weights fixed, not tuned. The corridor term is an expectation of a Gaussian kernel under the explicit Gaussian-mixture marginal and therefore has a closed form; $\mathcal L_\rho$ and $\mathcal L_\beta$ are deterministic nearest-neighbor/curvature penalties on the PWC parameters. Thus E1 and E2 optimize only $20$ variables, $(\rho_k,c_k)_{k=1}^{10}$, and do not optimize objective hyperparameters.

\paragraph{Gradients without inner Monte Carlo.} For the objective actually used in E1/E2, the loss is a composition of the PWC Riccati cascade, the algebraic Gaussian-mixture marginal, and closed-form Gaussian-kernel expectations. Reverse-mode differentiation therefore passes through finite-dimensional linear-algebra operations and requires no inner SDE simulation. The dominant analytic cascade scales as $O(Kd^3)$; E2 adds a finite average over the fixed source atoms. H1 deliberately performs no protocol optimization at all: it tests hand-crafted matrix schedules so that the high-dimensional experiment isolates geometry and scaling rather than hyperparameter search.
\end{revision}

\begin{revision}
\paragraph{What is optimized in this paper.} E1 and E2 jointly optimize the ten transverse offsets $\rho_k$ and ten transverse-stiffness parameters $c_k$ with $\sigma_t=0$, while the objective weights and the interval grid are fixed. H1 uses no optimization. This restricted experimental design is intentional: the aim is to expose and validate the analytic bridge backbone rather than to demonstrate a large hyperparameter-search procedure.

\paragraph{Beyond a deterministic source.} The exact density formula used in the theorem requires a deterministic source and a Gaussian-mixture target. For E2 we approximate the source by a fixed empirical measure and average deterministic-start conditional bridge quantities. A genuine continuous two-marginal problem generally requires an additional endpoint-scaling procedure (for example Sinkhorn/iterative proportional fitting), and we do not claim a closed-form reduction for that problem here.

\paragraph{Summary and bridge to experiments.} The next section therefore uses the analytic backbone in three deliberately controlled settings: exact deterministic-source protocol learning in E1; finite-empirical-source, sample-conditioned protocol learning in E2; and hand-crafted high-dimensional matrix protocols in H1.
\end{revision}

\section{Empirical demonstrations}
\label{sec:experiments}

\begin{revision}
The empirical sequence is intentionally simple-to-structured: E1 makes the analytic protocol-learning mechanism visually inspectable; E2 changes only the source treatment; H1 then tests the matrix-valued geometry in higher dimension. 
\end{revision}

\subsection{E1: 2D corridor with learned protocol}
\label{sec:E1}

\paragraph{Setup.} The first demonstration realizes the centerline-learning setting (item~1 in the list of \S\ref{sec:objective}) on a deliberately simple 2D geometry, so that the closed-form machinery, the density-level objective, and the resulting path improvement are all visually inspectable. The source is a delta at the origin, the target is the bimodal Gaussian mixture $p^{(\mathrm{tar})} = \tfrac12 \mathcal N\!\bigl((3, +0.5),\,\mathrm{diag}(0.06,0.05)\bigr)+ \tfrac12 \mathcal N\!\bigl((3, -0.5),\,\mathrm{diag}(0.06,0.05)\bigr)$, and a fixed \begin{revision}curved, single-arch corridor\end{revision} midline $m: [0,1] \to \R^2$ runs from the origin to $(3,0)$ and is generated by two opposing $\tanh$ transitions of amplitude $A = 0.7$ (the exact detrended formula is given in App.~\ref{app:caseB}). We use $K=10$ PWC intervals, $\sigma_t \equiv 0$, and $\kappa_t \equiv 1$. The trainable protocol consists of the per-interval transverse offset $\rho_k \in \R$ of the guide from the corridor midline ($\nu_k = m(s_k) + \rho_k\, n(s_k)$ with $n$ the local normal) and the transverse stiffness $\beta_k^{(\perp)} \in [2, 60]$ via a sigmoid reparameterization, with the longitudinal stiffness $\beta^{(\parallel)} = 0.2$ held fixed. \begin{revision}The baseline protocol places its ten PWC guide values on the corridor midline and uses isotropic $\beta=3$; it is therefore a weak-confinement baseline, not a straight-line guide.\end{revision}

\paragraph{Objective.} The corridor-alignment objective is the density-level functional
\begin{equation}
  \mathcal L_{\mathrm{corr}}(\Gamma)
  \;=\;
  \frac{1}{|\mathcal K|}\sum_{k \in \mathcal K}
  \Bigl(1 - \E_{X_{t_k} \sim p^*_{t_k}(\,\cdot\,;\Gamma)}
   \!\bigl[K_k(X_{t_k})\bigr]\Bigr),
  \quad
  K_k(x) = \exp\!\bigl(-\tfrac12 (x-m_k)^\top A_k (x-m_k)\bigr),
  \label{eq:E1_loss}
\end{equation}
where $A_k = Q_k\,\mathrm{diag}(\omega_\parallel^{-2}, \omega_\perp^{-2})\,Q_k^\top$ is the corridor kernel oriented by the local frame $Q_k = [t(s_k)\mid n(s_k)]$ with widths $(\omega_\parallel, \omega_\perp) = (0.8, 0.2)$, and the window $\mathcal K = \{k : s_k \le 0.8\}$ excludes the final intervals where the cloud is committing to the bimodal target. The expectation in Eq.~(\ref{eq:E1_loss}) is a closed-form Gaussian-mixture integral by Theorem~\ref{thm:closedform} --- a sum of $K_{\mathrm{tar}}$ Gaussian moment evaluations, no Monte-Carlo sampling. We add light $\rho$- and $c$-variation regularizers (App.~\ref{app:caseB}); none of the regularization terms requires sampling either.

\begin{revision}\paragraph{Optimization.} We initialize $\rho_k=0$ and $\beta_k^{(\perp)}=15$ and run 300 gradient steps with learning rate $3\times10^{-2}$. The $20$ protocol variables are the only optimized quantities; the objective weights in Eq.~(\ref{eq:actual_protocol_loss}) are fixed. Each step differentiates through the analytic Riccati/marginal calculation and uses no trajectory simulation. Throughout, terminal matching for the deterministic source remains exact by construction, so the optimization changes the path rather than the endpoint.\end{revision}

\paragraph{Result.}
\begin{figure}[htbp]
\centering
\includegraphics[width=0.98\linewidth]{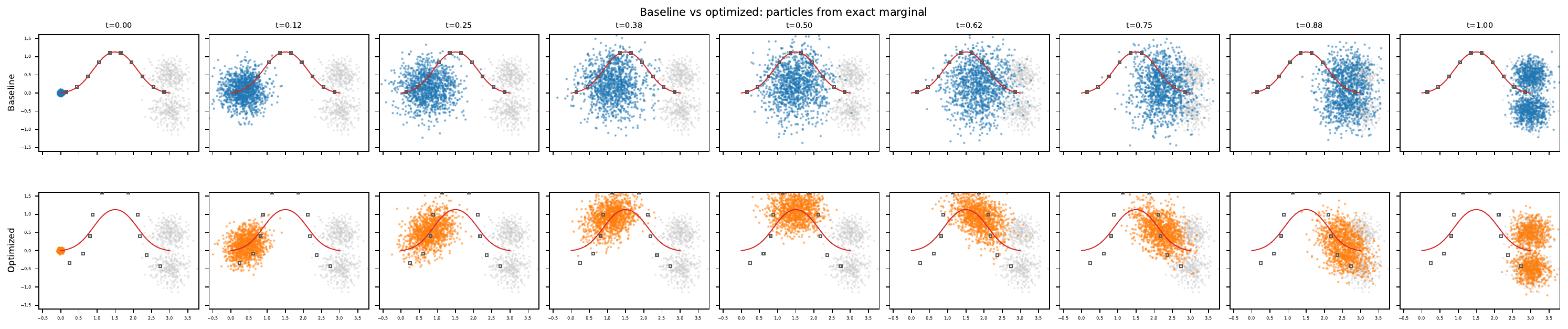}
\caption{\begin{revision}\textbf{E1: density-level corridor-protocol optimization.} Particles drawn from the closed-form marginal $p^*_t$ at nine times $t \in \{0, 0.12, 0.25, 0.38, 0.50, 0.62, 0.75, 0.88, 1\}$, under the baseline midline guide with isotropic $\beta = 3$ (\emph{top row}, blue) and under the density-level optimized guide with learned transverse offset $\rho_k$ and learned anisotropic $\beta_k^{(\perp)}$ (\emph{bottom row}, orange). Solid red curve: corridor midline. Open grey square markers: the ten PWC guide values $\nu_k=m(s_k)+\rho_k n(s_k)$; no continuous interpolation between these values is implied. Light grey points: target samples. The terminal panels ($t=1$) show that both protocols hit the bimodal target --- the LQ-GM-PID terminal law is exact by construction (Theorem~1). What the optimization buys is along the path: the optimized cloud follows the curved corridor much more closely, while the weak baseline cuts across its bend. Quantitatively the corridor-alignment loss Eq.~(\ref{eq:E1_loss}) drops from $0.7025$ to $0.4542$ (a $35.3\%$ reduction), with terminal mean-position error remaining at $\approx 0.003$ in both cases. The full study, including loss history and protocol-parameter plots, is in App.~\ref{app:caseB}.\end{revision}}
\label{fig:E1_comparison}
\end{figure}
\begin{revision}The result is shown in Fig.~\ref{fig:E1_comparison}. The baseline uses the same PWC midline guide values as the corridor but only weak isotropic confinement ($\beta=3$). The continuous-time bridge still has the prescribed terminal law by Theorem~\ref{thm:closedform}; the small numerical endpoint residuals shown in diagnostics come from evaluating at the boundary cutoff and from finite plotting samples. Along the path, however, the baseline cloud cuts the inside of the curved corridor, whereas the optimized anisotropic protocol keeps substantially more mass near the midline (Fig.~\ref{fig:E1_comparison}).\end{revision} After 300 gradient steps of density-level optimization, the cloud tracks the corridor: at every snapshot in the interior of the time window, the optimized $p^*_t$ has its mass concentrated near the midline, with the transverse spread compressed by the learned anisotropic $\beta_k^{(\perp)} \approx 50$ (compared to the baseline $\beta = 3$).

\begin{revision}
\paragraph{Why the optimized PWC guide values overshoot.} The optimized interval values $\nu_k$ lie beyond the corridor midline near the two bends, with peak transverse offset $\rho_k\approx0.55$. This is not a statement that $\nu_t$ is a smooth curve: the protocol is PWC, and the figures show the interval values as markers/steps. The overshoot itself has a dynamical interpretation. The center of $p_t^*$ depends on the full time history of the guide and lags rapidly changing guide values; placing $\nu_k$ ahead of the local bend compensates for this lag and keeps the probability mass near the corridor midline.
\end{revision}

\paragraph{Reading.} E1 establishes three concrete points used by the rest of the paper: (i)~the density-level objective Eq.~(\ref{eq:E1_loss}) is genuinely optimizable by autograd through the Riccati cascade, with no SDE simulation in the loop and no Monte-Carlo noise floor; (ii)~within the LQ-GM-PID class, optimization improves \emph{path quality} while leaving terminal exactness undisturbed --- the two are decoupled by construction; (iii)~the matrix-valued $\beta_t$ provides genuine anisotropic corridor confinement in 2D (a precursor to the high-dimensional corridor analogue of \S\ref{sec:H1}), and the combination of learned $\rho_k$ and learned $\beta_k^{(\perp)}$ is sufficient to produce the cleanly path-tracking dynamics of Fig.~\ref{fig:E1_comparison}. The full setup, loss-history curves, and protocol-parameter visualizations are in App.~\ref{app:caseB}; all numerical results in this section are reproducible via \path{experiments/exp_E1_corridor.ipynb}.

\subsection{\textcolor{blue}{E2: Sample-conditioned multi-entrance transport from a finite empirical source}}
\label{sec:E2}

\begin{revision}
\paragraph{Setup.} E2 keeps the E1 corridor, target, and $20$-parameter PWC protocol, but changes the source. We draw once from the bimodal Gaussian mixture
\begin{equation}
  p^{(\mathrm{in})}(x)=\tfrac12\mathcal N\bigl(x;(-0.3,+0.5),\sigma_0^2I_2\bigr)+\tfrac12\mathcal N\bigl(x;(-0.3,-0.5),\sigma_0^2I_2\bigr),\qquad \sigma_0=0.12,
  \label{eq:E2_source}
\end{equation}
and freeze $B=60$ sampled points $\{z^{(n)}\}_{n=1}^B$. The source modes are deliberately centered at $x=-0.3$, i.e. behind the corridor entrance at $x=0$.

\paragraph{What is analytic and what is approximated.} For every frozen source atom $z^{(n)}$, Theorem~\ref{thm:closedform} gives an exact deterministic-start conditional bridge with a $K_{\mathrm{tar}}$-component Gaussian-mixture marginal. We average these conditionals with equal source weights,
\begin{equation}
  p_t^{*,B}(x;\Gamma)=\frac1B\sum_{n=1}^{B}\sum_{k=1}^{K_{\mathrm{tar}}}\bar\pi^{(n)}_{k,t}\,
  \mathcal N\!\left(x;\mu^{(n)}_{k,t},(\Pi^{(n)}_{k,t})^{-1}\right),
  \label{eq:E2_marginal}
\end{equation}
where the component weights $\bar\pi^{(n)}_{k,t}$ are normalized separately for each deterministic start before the $1/B$ average. Thus $p_t^{*,B}$ is an analytic $120$-component mixture conditional on the fixed empirical source. It is an empirical approximation to transport from the continuous law in Eq.~(\ref{eq:E2_source}), not a claim that the continuous Gaussian-mixture-source bridge is closed form. In this experiment $\sigma_t=0$; the coordinate-shift shortcut used in App.~\ref{app:caseC} is restricted to that case.
\end{revision}

\begin{revision}
\paragraph{Density-level objective.} The E1 corridor loss is averaged over the fixed source atoms,
\begin{equation}
\mathcal L_{\mathrm{corr}}^{\mathrm{E2}}(\Gamma)=\frac1{|\mathcal K|}\sum_{k\in\mathcal K}\left(1-\frac1B\sum_{n=1}^B\mathbb E_{X_{t_k}\sim p^*_{t_k}(\cdot\mid z^{(n)},\Gamma)}[K_k(X_{t_k})]\right).
\label{eq:E2_loss}
\end{equation}
Each conditional expectation is an analytic Gaussian-mixture integral. The only random operation in the optimization is the one-time source draw; given that draw, the loss and gradient are deterministic. The regularization weights and optimizer settings are fixed exactly as in E1.
\end{revision}

\paragraph{Result.}
\begin{figure}[htbp]
\centering
\includegraphics[width=0.98\linewidth]{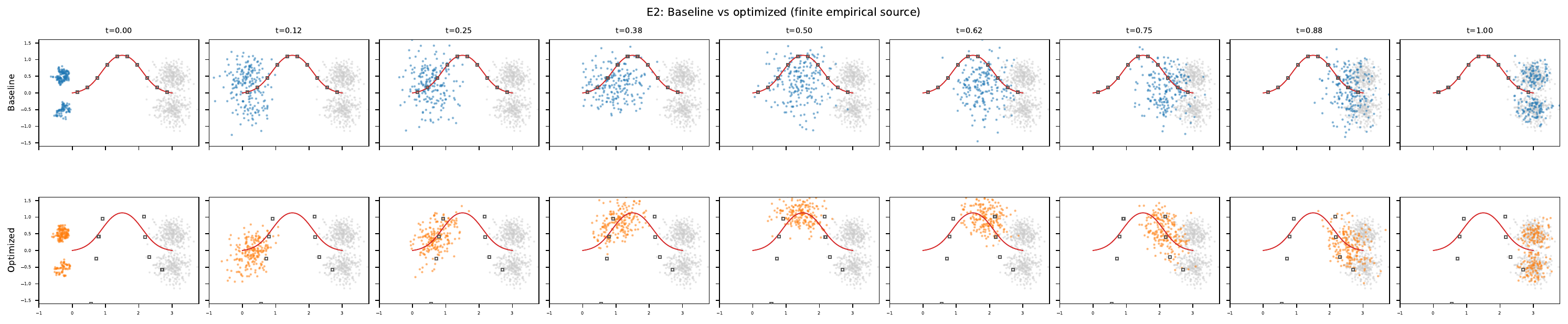}
\caption{\begin{revision}\textbf{E2: density-level corridor optimization from a finite empirical source.} Particles drawn from the finite-source analytic mixture $p_t^{*,B}$ at the same nine snapshot times as Fig.~\ref{fig:E1_comparison}, under the baseline midline guide with isotropic $\beta = 3$ (\emph{top row}, blue) and under the density-level optimized protocol (\emph{bottom row}, orange). Solid red curve: corridor midline. Open grey square markers in the bottom row: the optimized PWC guide values. Light grey points: target samples. The two entrance modes (visible at $t = 0$ as two clusters at $x \approx -0.3$, $y \approx \pm 0.5$) merge into the corridor neck, traverse the curved corridor, and resolve into the bimodal target. The corridor-alignment loss Eq.~(\ref{eq:E2_loss}) drops from $0.7733$ to $0.5074$ ($34.4\%$ reduction). Each deterministic-start conditional bridge has the prescribed terminal law; their equal-weight empirical mixture therefore does as well in the continuous-time construction. The full study, including loss history and protocol-parameter plots, is in App.~\ref{app:caseC}.\end{revision}}
\label{fig:E2_comparison}
\end{figure}
\begin{revision}
Fig.~\ref{fig:E2_comparison} shows the result. The finite-source corridor loss falls from $0.7733$ to $0.5074$ ($34.4\%$), close to the E1 reduction. The two entrance clusters are funneled into a narrow corridor-aligned sheet. The reported terminal moment residuals are numerical diagnostics of the finite-source conditional construction; exact terminal matching applies to each continuous-time deterministic-start conditional, while the continuous non-degenerate source itself is outside Theorem~\ref{thm:closedform}.
\end{revision}

\paragraph{The entrance-merging mechanism.}
\begin{figure}[htbp]
\centering
\includegraphics[width=0.95\linewidth]{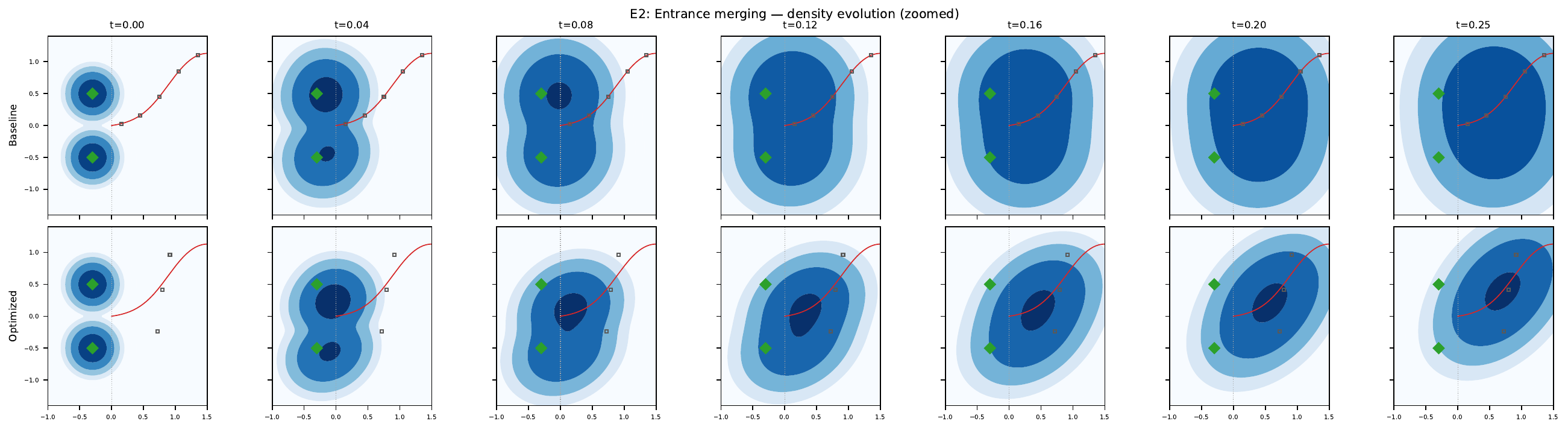}
\caption{\begin{revision}\textbf{E2 entrance-merging detail.} Finite-source analytic density $p_t^{*,B}$ at seven early times $t \in \{0, 0.04, 0.08, 0.12, 0.16, 0.20, 0.25\}$ under the baseline (\emph{top}) and optimized (\emph{bottom}) protocols, with color contours showing the $B \cdot K_{\mathrm{tar}} = 120$-component analytical mixture Eq.~(\ref{eq:E2_marginal}). Green diamonds mark the two entrance-mode means; solid red curve is the corridor midline; open grey squares in the bottom row are the PWC guide values; the vertical dotted line marks the corridor entrance $x=0$. Under the baseline, the two bumps merge through diffusive spread alone, producing a wide ``mushroom'' that extends transversely far beyond the corridor. Under the optimized protocol, the merge happens \emph{inside} the corridor: the optimized guide pulls each bump toward the corridor neck along the local normal, so the two streams enter the corridor as a single narrow sheet by $t = 0.12$.\end{revision}}
\label{fig:E2_merging}
\end{figure}
\begin{revision}
The early-time negative-$x$ mass is expected from the setup rather than a longitudinal overshoot: both source modes are centered at $x=-0.3$, behind the corridor entrance $x=0$, and retain source memory during the first part of the bridge. The early-time figure marks $x=0$ explicitly. The optimized PWC guide values mainly act in the transverse direction, pulling the two entrance modes toward the midline; by $t\simeq0.12$ the two streams have merged into a narrow sheet. Thus the early negative-$x$ bias should be read as finite-time transport from the deliberately displaced source, while the learned guide overshoot discussed in E1 refers to transverse offsets $\rho_k$.
\end{revision}

\begin{revision}
\paragraph{Reading.} E2 shows that the deterministic-start analytic backbone can be reused efficiently for a finite empirical source: the conditional losses remain analytic and the same $20$ protocol variables can be optimized. Its purpose is not to claim a closed-form Gaussian-mixture-to-Gaussian-mixture Schrödinger bridge, but to demonstrate a transparent sample-conditioned extension and the associated entrance-merging geometry.
\end{revision}

\subsection{H1: High-dimensional coarse-to-fine scaling}
\label{sec:H1}

\paragraph{Setup.} The third demonstration moves to high ambient dimension and high mode count to test whether the closed-form analytic backbone scales without optimization. \begin{revision}The source is a delta at the origin in $\R^d$. The target is stated explicitly as
\begin{equation}
p^{(\mathrm{tar})}(x)=\frac1M\sum_{b=1}^{B}\sum_{\ell=1}^{L}\mathcal N(x;\mu_{b\ell},\Sigma),\qquad
\mu_{b\ell}=\mu^{(T)}+\mu_b^{(B)}+\mu_\ell^{(L)},\quad M=BL,
\label{eq:H1_target_explicit}
\end{equation}
where the three offsets have disjoint trunk/branch/local support, $\mu^{(T)}=3e_1$, the branch-code amplitude is $1$, and the local-refinement amplitude is $0.35$. The target covariance is block diagonal,
\begin{equation}
\Sigma=\mathrm{diag}\bigl(0.10^2 I_{d_T},\,0.18^2 I_{d_B},\,0.25^2 I_{d_L}\bigr).
\label{eq:H1_covariance}
\end{equation}
The block sizes are $(d_T,d_B,d_L)=(1,2,1)$ for $d=4$, $(2,2,4)$ for $d=8$, and $(2,4,d-6)$ for $d\ge16$. Branch offsets are the sign-code directions specified in App.~\ref{app:H1}; local offsets occupy only the local block.\end{revision} A fast-then-hold trunk guide $\nu_t = \min(2t,1)\,\mu^{(\mathrm{trunk})}$ pulls the cloud onto the trunk endpoint by $t = 0.5$, and the branching is released over the back half of the time horizon. Three hand-crafted matrix-valued protocols $\beta_t \in \{B_0, B_1, B_2\}$ instantiate increasingly structured corridor geometries:
\begin{itemize}[leftmargin=1.5em]
\item \emph{B0 (isotropic baseline):} $\beta_t \equiv 2\,I_d$, identical on every block;
  
\item \emph{B1 (anisotropic corridor):} block-diagonal $\beta_t$ with loose trunk stiffness $\lambda_T = 0.5$ and tight orthogonal stiffness $\lambda_\perp = 4$, constant in time;

\item \emph{B2 (branch-release):} same as B1 except that the branch-block stiffness is held high ($\lambda_B = 6$) for $t < t_*= 0.5$ and released to $\tilde\lambda_B = 1$ for $t \ge t_*$, while the local block always uses $\lambda_L = 4$.
\end{itemize}
We sweep ambient dimension $d \in \{4, 8, 16, 32\}$ at fixed $M=8$ (\emph{H1-A}) and mode count $M \in \{2, 4, 8, 16\}$ at fixed $d=16$ (\emph{H1-B}); a representative scenario at $(d, M) = (16, 8)$ provides qualitative trunk-plane and principal-component-analysis (PCA) visualizations (\emph{H1-C}). The PWC partition uses $K=12$ uniform intervals so that $t_* = 0.5$ falls on a breakpoint. Diagnostics are collected from $B=1024$ Euler--Maruyama (EM) trajectories with $600$ time steps; subspace variance traces are reported both from the closed-form GMM marginal of Theorem~\ref{thm:closedform} and from the EM ensemble.

\paragraph{Result.}
\begin{figure}[htbp]
\centering
\includegraphics[width=0.99\linewidth]{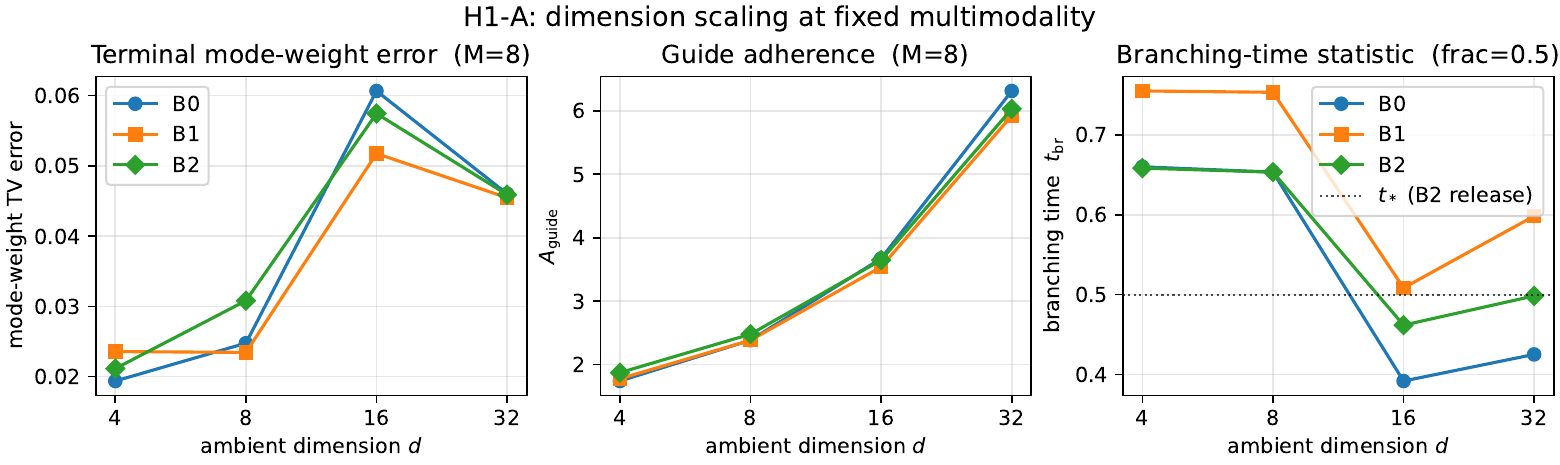}
\caption{\begin{revision}\textbf{H1-A: dimension scaling at fixed $M=8$.} Three diagnostics versus $d\in\{4,8,16,32\}$ for protocols B0, B1, and B2. \emph{Left:} terminal mode-weight total-variation error. \emph{Center:} integrated guide-adherence cost $A_{\mathrm{guide}}=\int_0^1\mathbb E\|X_t-\nu_t\|^2dt$. \emph{Right:} empirical branching time $t_{\mathrm{br}}$, with the design release $t_*=0.5$ shown by the dashed line. The deterministic-source analytic cascade has implementation-independent complexity $O(Kd^3)$, as discussed in the text.\end{revision}}
\label{fig:H1_main}
\end{figure}
The H1 results are summarized in Fig.~\ref{fig:H1_main}, which shows the three key diagnostics versus ambient dimension at fixed $M=8$. The mode-scaling sweep (versus $M$ at fixed $d=16$), the trunk-plane and PCA snapshots, and the closed-form vs.\ EM subspace-variance comparison are deferred to App.~\ref{app:H1}. The four scientific questions posed in the H1 design are answered as follows.

\paragraph{Q1: Does a simple fixed guide remain useful as $d$ grows?} Yes. At fixed $M = 8$, the terminal mode-weight total-variation (TV) error stays below $7\%$ across $d \in \{4,8,16,32\}$ for all three protocols (Fig.~\ref{fig:H1_main}, left). Note that this is a different quantity from the terminal mean error of E1/E2, which was always essentially zero by Theorem~\ref{thm:closedform}: the TV-error here measures empirical mode-weight allocation across the $M = 8$ components, which is sensitive to the relative timing of the branching events and the EM discretization, both of which are non-zero.

\paragraph{Q2: Does matrix-valued $\beta_t$ provide a meaningful high-dimensional corridor analogue?} Yes, most clearly in the path statistic. At $d=16$, B1 lowers $A_{\mathrm{guide}}$ from $3.67$ to $3.54$ and the terminal mode-weight TV error from $0.061$ to $0.052$; at $d=32$, it lowers $A_{\mathrm{guide}}$ from $6.31$ to $5.92$, while the TV errors of all three protocols are nearly indistinguishable ($\simeq0.045$--$0.046$). The improvement in guide adherence is not produced by tighter overall confinement --- B0 uses $\lambda=2$, larger than B1's trunk stiffness $\lambda_T=0.5$ --- but by the \emph{anisotropy}: loose along the trunk and tight perpendicular. The matrix-valued $\beta_t$ in B1 is the high-dimensional analogue of the corridor-aligned $\beta_k=Q_k\,\mathrm{diag}(\beta^{(\parallel)},\beta_k^{(\perp)})\,Q_k^\top$ of E1, with the trunk--branch--local block decomposition playing the role of the local frame.

\paragraph{Q3: Can branch-release scheduling control the branching time?} Yes (Fig.~\ref{fig:H1_main}, right). At $d=16$, the empirical branching times are $0.392$ (B0), $0.462$ (B2), and $0.508$ (B1); at $d=32$ they are $0.425$, $0.498$, and $0.598$, respectively. Thus B2 is the protocol whose branching statistic stays closest to the prescribed release $t_*=0.5$, while B0 branches earlier and B1 later. Its terminal TV error remains between those of B0 and B1 at both dimensions. This demonstrates that a piecewise-in-time branch-block stiffness can shift the onset of branching without degrading terminal mode allocation.

\paragraph{Q4: Does the method continue to separate coarse from fine resolution as $M$ grows?} Yes. The mode-scaling sweep at fixed $d = 16$ (Fig.~\ref{fig:H1B_mode_scaling} of App.~\ref{app:H1}) shows that the branching-time statistic depends sharply on the number of coarse branches $B$ and is insensitive to the local refinement count $L$: doubling $L$ at fixed $B$ leaves $t_{\mathrm{br}}$ unchanged at the reported time resolution, while $A_{\mathrm{guide}}$ changes by at most about $1\%$. The trunk--branch--local hierarchy is verifiable directly from the closed-form marginal: at every $(d, M, \mathrm{protocol})$ tested, the terminal block-variance trace satisfies $\mathrm{tr}(\mathrm{Var}_{\mathrm{trunk}}(X_1)) < \mathrm{tr}(\mathrm{Var}_{\mathrm{branch}}(X_1)) < \mathrm{tr}(\mathrm{Var}_{\mathrm{local}}(X_1))$, matching the target's design hierarchy. The closed-form variance traces of Theorem~\ref{thm:closedform} agree with the empirical EM estimate to plotting accuracy throughout the entire $t \in [0,1]$ window (Fig.~\ref{fig:H1A_subspace} of App.~\ref{app:H1}), providing simultaneous confirmation of both the analytic backbone and the EM integrator.

\paragraph{Reading.} H1 establishes that the LQ-GM-PID closed-form machinery scales to $d=32$ ambient dimensions and $M=16$ terminal modes \emph{without} optimization: hand-crafted matrix-valued protocols already exhibit the qualitative phenomena of structured high-dimensional transport (corridor adherence, controlled branching, coarse-to-fine hierarchy). Together, the optimized low-dimensional studies E1/E2 and the hand-crafted high-dimensional study H1 cover two complementary regimes of LQ-GM-PID: density-level protocol learning in visually inspectable geometries, and scalable matrix-protocol design in high dimension. Joint optimization of matrix-valued $\beta_t$ and $\sigma_t$ in high dimension is left to future work and is one of the natural extensions discussed in \S\ref{sec:discussion}.

\section{Discussion}
\label{sec:discussion}

This paper developed LQ-GM-PID as an analytically solvable reference class for bridge diffusions. The class is deliberately restricted: the guide potential is linear--quadratic, the state-dependent drift is linear, the protocol is piecewise constant in time, and the exact theorem uses a deterministic source and a Gaussian-mixture terminal distribution. Within this restriction, however, the bridge-diffusion problem becomes unusually transparent. The backward and forward Kolmogorov--Fokker--Planck equations reduce to a Riccati cascade; the optimal score is available as a closed-form softmax-weighted mixture of affine fields; the instantaneous marginal is itself an exact Gaussian mixture at every intermediate time; and protocol comparison can be carried out at the density level, without inner Monte-Carlo simulation or neural-network score fitting.

The empirical studies were designed to test whether this restricted class is still rich enough to reproduce the qualitative phenomena that motivate bridge-diffusion modeling. In the 2D corridor experiment, density-level optimization of the protocol improves path adherence while leaving terminal matching exact by construction. In E2, a fixed empirical sample from a Gaussian-mixture source is propagated through deterministic-start conditional bridges, producing a controlled entrance-merging mechanism inside the corridor. In the high-dimensional experiment, hand-crafted matrix-valued protocols already exhibit the intended coarse-to-fine behavior: anisotropic confinement, delayed branch release, and hierarchical variance growth in trunk, branch, and local subspaces. Thus the main message is not that LQ-GM-PID is a universal generative model, but that a substantial part of the geometry of bridge diffusion---path shaping, multimodal transport, and controlled branching---can be isolated in a class where the score, marginal, and protocol gradients remain explicit.

The benefit of this closed-form structure is fourfold. First, it gives a diagnostic baseline for neural bridge-diffusion methods. In the LQ-GM setting, the score field that a neural model would normally approximate is known exactly, making it possible to separate representation error, sampling error, and protocol-design error. \begin{revision}Second, it provides a low-variance environment for path-level objectives that reduce to explicit Gaussian-mixture expectations, such as the corridor-overlap loss used in E1/E2 and the subspace-moment diagnostics used in H1. More general functionals --- for example the squared norm of the full mixture score --- need not reduce to Gaussian second moments and may require numerical quadrature.\end{revision} This ultra-light evaluation is especially important in exploratory protocol design: when the objective is still being formulated, and the useful part of the parameter space is not yet known, fast autograd-supported scans can be more valuable than a highly expressive but expensive parameterization. Third, the protocol is interpretable. The matrices $\beta_t$ specify anisotropic stiffness, the centerline $\nu_t$ specifies the intended path geometry, the drift matrix $\sigma_t$ introduces state-dependent contraction, expansion, or rotation, and the diffusion schedule $\kappa_t$ controls stochastic spread. Fourth, the same analytic backbone provides a terminal look-up map $\widehat x_T(t;x)$: an inference-time summary of which terminal component or terminal region is implied by the current state. \begin{revision}This look-up capability is useful for biasing toward constrained terminal values or for diagnosing when a trajectory ensemble commits to a particular terminal mode.\end{revision}

The same closed-form restriction is also the main limitation of the present work. Gaussian mixtures are flexible but not universal in the operational sense required for high-dimensional data such as images, proteins, or empirical scientific datasets. The linear--quadratic structure is strong: it gives the Riccati solvability, but it excludes genuinely nonlinear guide potentials and nonlinear state-dependent drifts. The exact density-level protocol learning developed here is tied to a deterministic source and Gaussian-mixture target. For a non-degenerate source, E2 uses a finite empirical approximation; a genuine two-continuous-marginal bridge requires an additional endpoint scaling. Finally, the general matrix-$\sigma_t$ extension is only partially explored here. The analytic machinery supports linear state-dependent drift, but systematic protocol optimization with nonzero $\sigma_t$ raises stability and conditioning questions that deserve a separate treatment.

These limitations suggest several concrete directions. The first is a systematic $\sigma_t$ theory. In the present paper, $\sigma_t$ is included in the analytic backbone, but the experiments emphasize $\beta_t$ and $\nu_t$. A natural next step is to characterize which classes of matrix-valued $\sigma_t$ produce useful contraction, expansion, shear, or rotation without creating stiff or unstable Euler--Maruyama trajectories. A more ambitious extension is to make $\sigma_t$ stochastic, so that the state-dependent linear drift becomes a controlled random mixing mechanism rather than a deterministic protocol. This direction is attractive for applications where mixing itself is part of the phenomenon being modeled, including turbulence-inspired generative modeling and stochastic transport in high-dimensional physical systems.

The second direction is high-dimensional protocol learning. The H1 experiment used hand-crafted matrix-valued schedules, while E1 and E2 optimized low-dimensional corridor protocols. The next step is to combine these two regimes: learn structured matrix protocols in high dimension, using low-rank, block-diagonal, Kronecker, or graphical parameterizations of $\beta_t$ and $\sigma_t$ so that the number of protocol parameters remains controlled. This would preserve the interpretability of the LQ-GM protocol while moving beyond manually designed trunk--branch--local schedules.

The third direction is mean-field and broadcast guidance. LQ-GM-PID treats particles as conditionally independent once the protocol is fixed. In many applications, however, the useful control signal is population-level: samples should coordinate through empirical moments, density summaries, or shared task-level information. Mean-field PID provides one route to this setting, replacing independent path shaping by self-consistent or broadcast guidance. First steps in this direction, including connections to decision-oriented applications in energy systems, have already been taken in \citet{chertkov_mean-field_2026}. The closed-form LQ-GM backbone developed here is a natural starting point for such extensions, because it already exposes the marginal density and its moments explicitly.

The fourth direction is application-specific path sampling. Molecular transition paths, autonomous navigation, and decision-oriented sampling all care about the transient ensemble, not only the terminal distribution. These are precisely the settings in which a path-level protocol has meaning: one may want to avoid high-energy molecular configurations, keep a robotic ensemble inside a safety corridor, or delay branching decisions until sufficient information has accumulated. Gaussian mixtures provide a useful compromise for such applications: they are expressive enough to represent multimodal basins, entrances, exits, and intermediate route families, while remaining analytically tractable under the LQ-GM backbone. LQ-GM-PID will not solve these problems in its present restricted form, but it provides a controlled analytic laboratory in which path objectives, protocol parameterizations, and diagnostics can be developed before being transferred to neural or hybrid bridge-diffusion models.

Finally, the terminal look-up law introduced in App.~\ref{app:derivations}, \S\ref{appA:lookup_law}, suggests a sharper diagnostic of dynamic organization than the marginal path alone. The expectation $\widehat x_T(t;x)$ converts an intermediate state into its implied terminal endpoint, and the corresponding responsibilities quantify how strongly the trajectory has committed to one terminal mode. In H-PID and U-turn diffusion, such commitment can change abruptly in time and has been linked to dynamic phase-transition behavior \citep{behjoo_harmonic_2025,biroli_dynamical_2024,behjoo_u-turn_2025}. In LQ-GM-PID this diagnostic is explicit and differentiable, making it possible to study mode commitment, branch release, and order-parameter-like transitions directly at the density level.

In summary, LQ-GM-PID should be understood as an exactly solvable backbone for bridge diffusion rather than as a replacement for neural generative modeling. Its value is that it makes several usually hidden objects explicit: the score, the intermediate marginal, the effect of the protocol, the terminal look-up map, and the tradeoff between terminal matching and path shaping. This gives both a baseline for evaluating more expressive models and a constructive template for hybrid methods that combine analytic structure with learned components.

\paragraph{Acknowledgments:} The author thanks the University of Arizona start-up programme for financial support. This work was initiated during sabbatical visits to the University of Michigan Institute for Computational Discovery and Engineering, the International Centre for Theoretical Physics (ICTP), the Technische Universit\"at Ilmenau (Humboldt Fellowship), Lawrence Livermore National Laboratory (faculty mini-sabbatical program), KAIST Graduate School of AI and Ecole Normale Superior (Paris). Scientific engagement and encouragement from colleagues at all six institutions are gratefully acknowledged. 

\paragraph{Reproducibility:} In the interest of reproducibility, all Python/Jupyter code used to generate the figures and experiments reported in this paper is openly available at \href{https://github.com/mchertkov/LQ-GM-PID}{github.com/mchertkov/LQ-GM-PID}.

\paragraph{Use of LLM:} The author used large language models to assist with editing, code refactoring, and the organization of computational experiments. All mathematical derivations, scientific claims, and final code were independently checked, validated, and approved by the authors.


%

\appendix

\section{LQ-GM-PID derivations}
\label{app:derivations}

This appendix derives the closed-form analytic backbone of LQ-GM-PID referenced in \S\ref{sec:lqgmpid} of the main body. Specifically, it establishes: (i)~the controlled SDE, the Hamilton--Jacobi--Bellman equation, the Hopf--Cole transformation, and the optimal drift in ratio form (\S\ref{appA:general_pid}); (ii)~the Gaussian ansatz for the backward and forward Green functions on the LQ class (\S\ref{appA:gaussian_ansatz}); (iii)~the coefficient ODE system and its delta-function boundary conditions (\S\ref{appA:odes}); (iv)~the piecewise-constant matrix-exponential solutions of these ODEs and the PWC interface kicks (\S\ref{appA:pwc_closed_form}); (v)~the closed-form expressions for the optimal control and the marginal $p_t(x)$ as itself a Gaussian mixture, when the target is a Gaussian mixture and the source is a delta (\S\ref{appA:gmm_target_explicit}); and \begin{revision}(vi)~the finite-empirical-source, sample-conditioned reduction used in E2, including the zero-$\sigma$ coordinate-shift shortcut (\S\ref{appA:gmm_initial}).\end{revision}

Notation conventions follow the main body: the time horizon is $[0,T]$, with $T=1$ in all numerical experiments; the state lives in $\R^d$; the protocol is $\Gamma_t=(\beta_t,\nu_t,\sigma_t,\kappa_t)$; the diffusion coefficient $\kappa_t>0$ is a scalar function of time. The target is denoted $p^{(\mathrm{tar})}$, the source $p^{(\mathrm{in})}$.

\subsection{General Path Integral Diffusion formulas}
\label{appA:general_pid}

Continuing the line of work on Harmonic Path Integral Diffusion (H-PID) \citep{behjoo_harmonic_2025,chertkov_adaptive_2026,chertkov_generative_2025}, we study the (controlled) stochastic process defined and the cost-to-go (value) function, defined in Eqs.~(in \ref{eq:sde_lqgmpid},\ref{eq:cost_to_go}) which we reproduce here for convenience
\begin{align}
  J_t(x)
  &= \inf_{u_{t\to T}}
     \E\Bigl[
        \int_t^{T}
        \Bigl(\tfrac{1}{2\kappa_{t'}} \|u_{t'}\|^2 + V_{t'}(X_{t'})\Bigr)\dd t'
        + \phi(X_T)
        \,\Big|\, X_t=x
     \Bigr],
  \label{appA:eq:cost_to_go}\\
  \dd X_t &= \bigl(f_t(X_t)+u_t(X_t)\bigr)\dd t + \sqrt{\kappa_t}\,\dd W_t,\quad X_0\sim p^{(\mathrm{in})},
  \quad X_t\in\R^d.
  \label{appA:eq:sde_controlled}
\end{align}

\paragraph{Hamilton--Jacobi--Bellman and Hopf--Cole.} The Hamilton--Jacobi--Bellman (HJB) equation for the cost-to-go is
\begin{align}
  -\partial_t J_t(x)
  &= V_t(x) + f_t(x)\cdot \n J_t(x)
     + \tfrac{\kappa_t}{2} \Del J_t(x)
     - \tfrac{\kappa_t}{2} \|\n J_t(x)\|^2,
  \qquad J_T(x)=\phi(x),
  \label{appA:eq:hjb}
\end{align}
with optimal control (score function)
\begin{equation}
  u_t^*(x) = -\kappa_t\,\n J_t(x) = \kappa_t\,\n \log \psi_t(x),
  \qquad \psi_t(x)\doteq e^{-J_t(x)},
  \label{appA:eq:u_star}
\end{equation}
where all gradients/Laplacians below are with respect to the \emph{spatial} argument $x$, and in $\partial_t$ we keep $x$ fixed.

Under the Hopf--Cole transform, $\psi_t=e^{-J_t}$ solves the linear ``lin--HJB'' PDE
\begin{align}
  -\partial_t \psi_t(x) + V_t(x)\psi_t(x)
  &= f_t(x)\cdot \n \psi_t(x) + \tfrac{\kappa_t}{2} \Del \psi_t(x),
  \qquad \psi_T(x)=e^{-\phi(x)}.
  \label{appA:eq:lin_hjb}
\end{align}

\paragraph{Green functions (backward/forward).}
Define the backward and forward Green functions, $G_t^{(-)}(x\mid y)$ and $G_t^{(+)}(x\mid y)$, as solutions of
\begin{align}
  &t\in[T\to 0]:\quad
  -\partial_t G_t^{(-)}(x\mid y) + V_t(x)G_t^{(-)}(x\mid y)
  = f_t(x)\cdot \n_x G_t^{(-)}(x\mid y) + \tfrac{\kappa_t}{2} \Del_x G_t^{(-)}(x\mid y),
  \label{appA:eq:GF_minus}
  \\
  &t\in[0\to T]:\quad
  \partial_t G_t^{(+)}(x\mid y) + V_t(x)G_t^{(+)}(x\mid y)
  = -\n_x\!\cdot\!\bigl(f_t(x)G_t^{(+)}(x\mid y)\bigr) + \tfrac{\kappa_t}{2} \Del_x G_t^{(+)}(x\mid y),
  \label{appA:eq:GF_plus}
\end{align}
with $\,G_T^{(-)}(x\mid y)=\delta(x-y)\,$ and $\;G_0^{(+)}(x\mid y)=\delta(x-y)$.

\paragraph{Convolutions and optimal marginals.} By linearity, the lin--HJB solution is
\begin{equation}
  \psi_t(x) = \int_{\R^d} e^{-\phi(y)}\,G_t^{(-)}(x\mid y)\,\dd y.
  \label{appA:eq:psi_convolution}
\end{equation}

\paragraph{Controlled Kolmogorov--Fokker--Planck and factorization.} Under a (feedback) control $u_t(x)$, the marginal density $p_t(x)$ of Eq.~(\ref{appA:eq:sde_controlled}) satisfies the controlled Kolmogorov--Fokker--Planck (KFP) equation
\begin{align}
  \partial_t p_t(x)
  &= -\nabla\!\cdot\!\Bigl((f_t(x)+u_t(x))\,p_t(x)\Bigr)
     + \tfrac{\kappa_t}{2} \Delta p_t(x),
  \qquad p_0(x)=p^{(0)}(x).
  \label{appA:eq:kfp_controlled}
\end{align}
For the optimal control $u_t^*(x)=\kappa_t\nabla \log \psi_t(x)$ from Eq.~(\ref{appA:eq:u_star}), the optimal density $p_t^*(x)$ solves Eq.~(\ref{appA:eq:kfp_controlled}) with $u_t=u_t^*$ and admits the factorization
\begin{equation}
  p_t^*(x)
  = \frac{\varphi_t(x)\,\psi_t(x)}{\int_{\R^d}\varphi_t(x')\psi_t(x')\,\dd x'},
  \label{appA:eq:p_star_product}
\end{equation}
where $\psi_t$ is the backward lin--HJB solution Eq.~(\ref{appA:eq:lin_hjb}), and $\varphi_t$ solves the adjoint (forward) linear equation
\begin{align}
  \partial_t \varphi_t(x) + V_t(x)\varphi_t(x)
  &= -\nabla\!\cdot\!\bigl(f_t(x)\varphi_t(x)\bigr)
     + \tfrac{\kappa_t}{2} \Delta \varphi_t(x),
  \label{appA:eq:varphi_forward}
\end{align}
with an initial condition determined by $p^{(0)}$ (e.g., $\varphi_0=\delta(\cdot-x_0)$ for deterministic start). In terms of the Green functions Eqs.~(\ref{appA:eq:GF_minus}--\ref{appA:eq:GF_plus}),
\begin{align}\label{appA:eq:psi+phi}
\psi_t(x)=\int e^{-\phi(y)}G_t^{(-)}(x\mid y)\,\dd y,\qquad
\varphi_t(x)=\int G_t^{(+)}(x\mid y)\, \varphi_0(y)\,\dd y.
\end{align}

\paragraph{Ratio form of the optimal control.} Eliminating $\phi$ by combining Eqs.~(\ref{appA:eq:u_star}--\ref{appA:eq:p_star_product}), and~Eq.~(\ref{appA:eq:psi+phi}), the forward and backward messages, the marginal density, and the optimal control are expressed via the Green functions and the prescribed source/target densities:
\begin{align} \nonumber 
  \varphi_t(x) &= \int_{\R^d} G_t^{(+)}(x\mid z)\,p^{(\mathrm{in})}(z)\,\dd z,\quad
  \psi_t(x)  \propto\ \int_{\R^d}
    \frac{p^{(\mathrm{tar})}(y)}{\varphi_T(y)}\, G_t^{(-)}(x\mid y)\,\dd y,
  \\ \label{appA:eq:p_star-via-G}
  p_t^{(*)}(x) & \propto \varphi_t(x)\,\psi_t(x)
  \\ \nonumber & \propto
  \left(\int_{\R^d} G_t^{(+)}(x\mid z)\,p^{(\mathrm{in})}(z)\,\dd z\right)
  \left(\int_{\R^d}
    \frac{p^{(\mathrm{tar})}(y) G_t^{(-)}(x\mid y)}{
          \int_{\R^d} G_T^{(+)}(y\mid z)\,p^{(\mathrm{in})}(z)\,\dd z}\,\dd y\right),
  \\ \label{appA:eq:u_star-via-G}
  u_t^*(x) &= \kappa_t \nabla_x \log \psi_t(x)
  =\kappa_t \nabla_x \log\!\left(\int_{\R^d}
    \frac{p^{(\mathrm{tar})}(y) G_t^{(-)}(x\mid y)}{
          \int_{\R^d} G_T^{(+)}(y\mid z)\,p^{(\mathrm{in})}(z)\,\dd z}\,\dd y\right).
\end{align}
Eq.~(\ref{appA:eq:u_star-via-G}) is the \emph{ratio form} of the optimal control: the score is the spatial gradient of the logarithm of a backward integral of the target density against $G^{(-)}$, normalised by the forward propagator at the terminal time. The remaining subsections specialize this ratio form to the LQ-GM class, where every integral becomes Gaussian and closed-form expressions emerge.

\subsection{Gaussian ansatz for linear drift and quadratic potential}
\label{appA:gaussian_ansatz}

We now specialize to
\begin{equation}
  f_t(x)=\sigma_t x,\qquad
  V_t(x)= \tfrac12(x-\nu_t)^\top \beta_t (x-\nu_t),
  \label{appA:eq:linear_drift_quadratic_potential_user}
\end{equation}
where $\beta_t\in\R^{d\times d}$ is symmetric positive definite and
$\sigma_t\in\R^{d\times d}$ is arbitrary (time-dependent).

\subsubsection{Gaussian ansatz for Green functions}

We postulate the Gaussian--exponential form (for each choice of the
$\pm$ branch)
\begin{align}
  G_t^{(\pm)}(x\mid y)
  &=
  \exp\Bigl(
      -\tfrac12 x^\top A_t^{(\pm)} x
      + x^\top B_t^{(\pm)} y
      -\tfrac12 y^\top C_t^{(\pm)} y
      + (\theta_{x;t}^{(\pm)})^\top x
      + (\theta_{y;t}^{(\pm)})^\top y
      + \zeta_t^{(\pm)}
    \Bigr),
  \label{appA:eq:pm_ansatz_clean}
\end{align}
where for each $t$:
\begin{itemize}
  \item $A_t^{(\pm)}$ and $C_t^{(\pm)}$ are symmetric ($d\times d$) matrices;
  \item $B_t^{(\pm)}$ is a general ($d\times d$) matrix;
  \item $\theta_{x;t}^{(\pm)},\theta_{y;t}^{(\pm)}\in\R^d$;
  \item $\zeta_t^{(\pm)}\in\R$ absorbs normalizing terms (including the
    singular prefactor needed for the delta-function boundary conditions).
\end{itemize}

\subsubsection{Useful differential identities}

For any $G(x)=\exp(\ell(x))$ with $\ell$ smooth,
\begin{equation}
  \frac{\nabla_x G}{G} = \nabla_x \ell,
  \qquad
  \frac{\Delta_x G}{G} = \|\nabla_x \ell\|^2 + \tr(\nabla_x^2 \ell).
  \label{appA:eq:exp_identities_clean}
\end{equation}
For Eq.~(\ref{appA:eq:pm_ansatz_clean}), define
\begin{equation}
  \ell_t^{(\pm)}(x\mid y)
  \doteq \log G_t^{(\pm)}(x\mid y).
\end{equation}
Then
\begin{align}
  \nabla_x \ell_t^{(\pm)}(x\mid y)
  &= -A_t^{(\pm)}x + B_t^{(\pm)}y + \theta_{x;t}^{(\pm)},
  \label{appA:eq:grad_logG_pm}
  \\
  \nabla_x^2 \ell_t^{(\pm)}(x\mid y)
  &= -A_t^{(\pm)}.
  \label{appA:eq:hess_logG_pm}
\end{align}

\subsection{Coefficient ODEs and boundary conditions}
\label{appA:odes}

Inserting Eq.~(\ref{appA:eq:pm_ansatz_clean}) into Eq.~(\ref{appA:eq:GF_minus})
(resp.\ Eq.~(\ref{appA:eq:GF_plus})) and matching coefficients of the
resulting polynomial identity in $(x,y)$ yields a closed ODE system for
the coefficient families
\[
A_t^{(\pm)},\,B_t^{(\pm)},\,C_t^{(\pm)},\,
\theta_{x;t}^{(\pm)},\,\theta_{y;t}^{(\pm)},\,\zeta_t^{(\pm)}.
\]
Below we record the resulting evolution equations.

\subsubsection{Backward branch}

For the backward Green function Eq.~(\ref{appA:eq:GF_minus}), the
coefficients satisfy:

\paragraph*{Quadratic (matrix) coefficients.}
\begin{align}
  \dot A_t^{(-)}
  &=
  \kappa_t\,A_t^{(-)}A_t^{(-)}
  - \Bigl(\sigma_t^\top A_t^{(-)} + A_t^{(-)}\sigma_t\Bigr)
  - \beta_t,
  \label{appA:eq:ode_A_minus_clean}
  \\
  \dot B_t^{(-)}
  &=
  \kappa_t\,A_t^{(-)} B_t^{(-)} - \sigma_t^\top B_t^{(-)},
  \label{appA:eq:ode_B_minus_clean}
  \\
  \dot C_t^{(-)}
  &=
  \kappa_t\,(B_t^{(-)})^\top B_t^{(-)}.
  \label{appA:eq:ode_C_minus_clean}
\end{align}

\paragraph*{Linear (vector) coefficients.}
\begin{align}
  \dot \theta_{x;t}^{(-)}
  &=
  \Bigl(\kappa_t\,A_t^{(-)}-\sigma_t^\top\Bigr)\theta_{x;t}^{(-)}
  - \beta_t \nu_t,
  \label{appA:eq:ode_thetax_minus_clean}
  \\
  \dot \theta_{y;t}^{(-)}
  &=
  - \kappa_t\,(B_t^{(-)})^\top \theta_{x;t}^{(-)}.
  \label{appA:eq:ode_thetay_minus_clean}
\end{align}

\paragraph*{Scalar normalizer.}
\begin{equation}
  \dot \zeta_t^{(-)}
  =
  \tfrac{\kappa_t}{2} \tr\!\bigl(A_t^{(-)}\bigr)
  + \tfrac12 \nu_t^\top \beta_t \nu_t
  - \tfrac{\kappa_t}{2} \|\theta_{x;t}^{(-)}\|^2.
  \label{appA:eq:ode_zeta_minus_clean}
\end{equation}

\subsubsection{Forward branch}

For the forward Green function Eq.~(\ref{appA:eq:GF_plus}), the
coefficients satisfy:

\paragraph*{Quadratic (matrix) coefficients.}
\begin{align}
  \dot A_t^{(+)}
  &=
  \beta_t
  - \Bigl(\sigma_t^\top A_t^{(+)} + A_t^{(+)}\sigma_t\Bigr)
  - \kappa_t\,A_t^{(+)}A_t^{(+)},
  \label{appA:eq:ode_A_plus_clean}
  \\
  \dot B_t^{(+)}
  &=
  -\Bigl(\kappa_t\,A_t^{(+)}+\sigma_t^\top\Bigr) B_t^{(+)},
  \label{appA:eq:ode_B_plus_clean}
  \\
  \dot C_t^{(+)}
  &=
  - \kappa_t\,(B_t^{(+)})^\top B_t^{(+)}.
  \label{appA:eq:ode_C_plus_clean}
\end{align}

\paragraph*{Linear (vector) coefficients.}
\begin{align}
  \dot \theta_{x;t}^{(+)}
  &=
  -\Bigl(\sigma_t^\top + \kappa_t\,A_t^{(+)}\Bigr)\theta_{x;t}^{(+)}
  + \beta_t \nu_t,
  \label{appA:eq:ode_thetax_plus_clean}
  \\
  \dot \theta_{y;t}^{(+)}
  &=
  \kappa_t\,(B_t^{(+)})^\top \theta_{x;t}^{(+)}.
  \label{appA:eq:ode_thetay_plus_clean}
\end{align}

\paragraph*{Scalar normalizer.}
\begin{equation}
  \dot \zeta_t^{(+)}
  =
  -\tr(\sigma_t)
  - \tfrac{\kappa_t}{2} \tr\!\bigl(A_t^{(+)}\bigr)
  + \tfrac{\kappa_t}{2} \|\theta_{x;t}^{(+)}\|^2
  - \tfrac12 \nu_t^\top \beta_t \nu_t.
  \label{appA:eq:ode_zeta_plus_clean}
\end{equation}

\subsubsection{Boundary conditions (delta-function limits)}

The Green functions are pinned by
\[
G_T^{(-)}(x\mid y)=\delta(x-y),\qquad G_0^{(+)}(x\mid y)=\delta(x-y).
\]
In the Gaussian representation Eq.~(\ref{appA:eq:pm_ansatz_clean}), these
are enforced as singular ``heat-kernel'' limits.

\paragraph*{Backward delta at $t\uparrow T$.}
As $t\uparrow T$, the dominant quadratic form must collapse onto
$(x-y)$:
\begin{align}
  A_t^{(-)} \sim \frac{1}{\kappa_T(T-t)}I,\qquad
  B_t^{(-)} \sim \frac{1}{\kappa_T(T-t)}I,\qquad
  C_t^{(-)} \sim \frac{1}{\kappa_T(T-t)}I,
  \label{appA:eq:bc_minus_ABC_clean}
\end{align}
and the linear terms must vanish at the delta boundary:
\begin{equation}
  \lim_{t\uparrow T}\theta_{x;t}^{(-)} = 0,\qquad
  \lim_{t\uparrow T}\theta_{y;t}^{(-)} = 0.
  \label{appA:eq:bc_minus_theta_clean}
\end{equation}

\paragraph*{Forward delta at $t\downarrow 0$.}
As $t\downarrow 0$,
\begin{align}
  A_t^{(+)} \sim \frac{1}{\kappa_0\,t}I,\qquad
  B_t^{(+)} \sim \frac{1}{\kappa_0\,t}I,\qquad
  C_t^{(+)} \sim \frac{1}{\kappa_0\,t}I,
  \label{appA:eq:bc_plus_ABC_clean}
\end{align}
and the linear terms must vanish at the delta boundary:
\begin{equation}
  \lim_{t\downarrow 0}\theta_{x;t}^{(+)} = 0,\qquad
  \lim_{t\downarrow 0}\theta_{y;t}^{(+)} = 0.
  \label{appA:eq:bc_plus_theta_clean}
\end{equation}

\subsection{Piecewise-constant protocols: closed-form coefficient updates}
\label{appA:pwc_closed_form}

We now assume a piecewise-constant (PWC) protocol on a grid
\[
0=t_0<t_1<\cdots<t_K=T,
\]
where, on each interval $I_k=[t_k,t_{k+1}]$,
\begin{equation}
  \beta_t=\beta_k,\qquad \nu_t=\nu_k,\qquad \sigma_t=\sigma_k,
  \qquad \kappa_t=\kappa_k,
  \qquad t\in I_k,
  \label{appA:eq:pwc_params}
\end{equation}
with $\beta_k=\beta_k^\top\succ 0$, $\kappa_k>0$, and $\sigma_k$ arbitrary.

\paragraph*{Practical note on endpoint singularities.}
The delta boundary conditions at $t=0$ and $t=T$ are singular. The
closed-form PWC construction below imposes the \emph{asymptotic}
behavior Eq.~(\ref{appA:eq:bc_minus_ABC_clean}--\ref{appA:eq:bc_plus_theta_clean}).
In applications (e.g.\ simulating the SDE with $u_t^*$), one simply
avoids querying coefficients too close to the endpoints by working on
$t\in[\varepsilon,T-\varepsilon]$ with a small $\varepsilon>0$.

Throughout this section we propagate coefficients on a single interval
$I_k$. We write $h_k\doteq t_{k+1}-t_k$ and use the local time
$\tau\in[0,h_k]$.

\subsubsection{Backward branch: explicit update on an interval}

Recall the backward-branch ODEs
Eq.~(\ref{appA:eq:ode_A_minus_clean}--\ref{appA:eq:ode_zeta_minus_clean}).
On $I_k$ define the $2d\times 2d$ constant ``Hamiltonian'' matrix
\begin{equation}
  H_k^{(-)}
  \doteq
  \begin{pmatrix}
    \sigma_k & -\kappa_k I\\
    -\beta_k & -\sigma_k^\top
  \end{pmatrix}.
  \label{appA:eq:Hminus_def}
\end{equation}
Let the block matrix exponential be
\begin{equation}
  \Phi_k^{(-)}(\tau)\doteq \exp\!\bigl(-H_k^{(-)}\,\tau\bigr)
  =
  \begin{pmatrix}
    \Phi^{(-)}_{11}(\tau) & \Phi^{(-)}_{12}(\tau)\\
    \Phi^{(-)}_{21}(\tau) & \Phi^{(-)}_{22}(\tau)
  \end{pmatrix},
  \qquad \tau\in[0,h_k].
  \label{appA:eq:Hminus_expm_blocks}
\end{equation}

\paragraph*{Initialization and induction across intervals.}
The backward recursion is initialized by the delta asymptotics at
$t\uparrow T$,
Eq.~(\ref{appA:eq:bc_minus_ABC_clean}--\ref{appA:eq:bc_minus_theta_clean}).
In practice, one evaluates the coefficients on $t\le T-\varepsilon$ and
starts the recursion at a time $t_\star=T-\varepsilon$ with large values
consistent with the asymptotic (e.g.\
$A_{t_\star}^{(-)}\approx (\kappa_K\varepsilon)^{-1}I$,
$B_{t_\star}^{(-)}\approx (\kappa_K\varepsilon)^{-1}I$,
$C_{t_\star}^{(-)}\approx (\kappa_K\varepsilon)^{-1}I$ and
$\theta_{x;t_\star}^{(-)}=\theta_{y;t_\star}^{(-)}=0$).
After computing coefficients on an interval, enforce continuity at
each grid point and use the left-end values as the terminal data for
the next (earlier) interval.

\paragraph*{Step 1: quadratic coefficients (explicit).}
Fix terminal values at $t_{k+1}$ (right endpoint of the interval):
\[
A_{k+1}^{(-)} \doteq A^{(-)}_{t_{k+1}},\quad
B_{k+1}^{(-)} \doteq B^{(-)}_{t_{k+1}},\quad
C_{k+1}^{(-)} \doteq C^{(-)}_{t_{k+1}}.
\]
Define
\begin{align}
  X_k^{(-)}(\tau) &\doteq \Phi^{(-)}_{11}(\tau) + \Phi^{(-)}_{12}(\tau)\,A_{k+1}^{(-)},
  \label{appA:eq:Xminus_def}
  \\
  Y_k^{(-)}(\tau) &\doteq \Phi^{(-)}_{21}(\tau) + \Phi^{(-)}_{22}(\tau)\,A_{k+1}^{(-)}.
  \label{appA:eq:Yminus_def}
\end{align}
Then for $t=t_{k+1}-\tau$,
\begin{align}
  A_t^{(-)} &= Y_k^{(-)}(\tau)\,\bigl(X_k^{(-)}(\tau)\bigr)^{-1},
  \label{appA:eq:pwc_A_minus}
  \\
  B_t^{(-)} &= \bigl(X_k^{(-)}(\tau)\bigr)^{-\top}\,B_{k+1}^{(-)}.
  \label{appA:eq:pwc_B_minus}
\end{align}
Moreover, $C_t^{(-)}$ can be written in closed form without time
integrals by augmenting the linear flow. Introduce the $3d\times 3d$
augmented matrix
\begin{equation}
  \mathcal H_k^{(-)}
  \doteq
  \begin{pmatrix}
    \sigma_k & -\kappa_k I & 0\\
    -\beta_k & -\sigma_k^\top & 0\\
    0 & -B_{k+1}^{(-)\top} & 0
  \end{pmatrix},
  \label{appA:eq:Hminus_aug_C}
\end{equation}
and let
\begin{equation}
  \exp\!\bigl(- \mathcal H_k^{(-)}\tau\bigr)
  =
  \begin{pmatrix}
    \ast & \ast & \ast\\
    \ast & \ast & \ast\\
    \Gamma^{(-)}_1(\tau) & \Gamma^{(-)}_2(\tau) & I
  \end{pmatrix},
  \label{appA:eq:Hminus_aug_blocks}
\end{equation}
where $\Gamma^{(-)}_1(\tau),\Gamma^{(-)}_2(\tau)\in\R^{d\times d}$ denote
the bottom-left blocks. Then
\begin{equation}
  C_t^{(-)}
  = C_{k+1}^{(-)} - \Gamma^{(-)}_2(\tau)\,\bigl(X_k^{(-)}(\tau)\bigr)^{-1}\,B_{k+1}^{(-)}.
  \label{appA:eq:pwc_C_minus}
\end{equation}

\paragraph*{Step 2: linear coefficients (explicit).}
Fix terminal values
\[
\theta_{x;k+1}^{(-)} \doteq \theta_{x;t_{k+1}}^{(-)},\qquad
\theta_{y;k+1}^{(-)} \doteq \theta_{y;t_{k+1}}^{(-)}.
\]
Define the constant drive vector on $I_k$:
\begin{equation}
  g_k \doteq \beta_k \nu_k \in\R^d.
  \label{appA:eq:gk_def}
\end{equation}
Define the $(2d+1)\times(2d+1)$ augmented matrix
\begin{equation}
  \widehat H_k^{(-)}
  \doteq
  \begin{pmatrix}
    \sigma_k & -\kappa_k I & 0\\
    -\beta_k & -\sigma_k^\top & 0\\
    g_k^\top & 0 & 0
  \end{pmatrix},
  \label{appA:eq:Hminus_aug_theta}
\end{equation}
and its exponential blocks
\begin{equation}
  \exp\!\bigl(-\widehat H_k^{(-)}\tau\bigr)
  =
  \begin{pmatrix}
    \Phi^{(-)}_{11}(\tau) & \Phi^{(-)}_{12}(\tau) & 0\\
    \Phi^{(-)}_{21}(\tau) & \Phi^{(-)}_{22}(\tau) & 0\\
    \widetilde\Gamma^{(-)}_1(\tau) & \widetilde\Gamma^{(-)}_2(\tau) & 1
  \end{pmatrix},
  \label{appA:eq:Hminus_aug_theta_blocks}
\end{equation}
where $\widetilde\Gamma^{(-)}_1(\tau),\widetilde\Gamma^{(-)}_2(\tau)\in\R^{1\times d}$
are the bottom-left blocks, and the upper $2d\times 2d$ corner equals
$\Phi_k^{(-)}(\tau)$ from Eq.~(\ref{appA:eq:Hminus_expm_blocks}). Then
\begin{equation}
  \theta_{x;t}^{(-)}
  =
  \bigl(X_k^{(-)}(\tau)\bigr)^{-1}
  \Bigl(
    \widetilde\Gamma^{(-)}_1(\tau)
    + \widetilde\Gamma^{(-)}_2(\tau)\,A_{k+1}^{(-)}
    + \theta_{x;k+1}^{(-)}
  \Bigr),
  \qquad t=t_{k+1}-\tau,
  \label{appA:eq:pwc_thetax_minus}
\end{equation}
where $X_k^{(-)}(\tau)$ is the same matrix as in
Eq.~(\ref{appA:eq:Xminus_def}). Finally,
\begin{equation}
  \theta_{y;t}^{(-)}
  =
  \theta_{y;k+1}^{(-)}
  + \kappa_k\int_0^{\tau} \bigl(B^{(-)}(s)\bigr)^\top\,\theta_{x}^{(-)}(s)\,\dd s,
  \label{appA:eq:pwc_thetay_minus}
\end{equation}
where $B^{(-)}(s)=\bigl(X_k^{(-)}(s)\bigr)^{-\top}B_{k+1}^{(-)}$
(Eq.~(\ref{appA:eq:pwc_B_minus}) and $\theta_{x}^{(-)}(s)$ is given by
Eq.~(\ref{appA:eq:pwc_thetax_minus}) evaluated at elapsed time $s$.

\paragraph*{Isotropic case $\sigma_k=0$, $\beta_k=b_k I$.}
With $w_k=\sqrt{\kappa_k\, b_k}$, the matrix-coefficient ODEs decouple
into $d$ identical scalar problems. One finds
\begin{equation}
  \theta_{x}^{(-)}(\tau)
  = \frac{g_k}{w_k}\tanh\!\Bigl(\frac{w_k\tau}{2}\Bigr),
  \label{appA:eq:thetax_isotropic}
\end{equation}
and substituting into the integrand $\kappa_k\,B(s)^\top\theta_x(s) =
\kappa_k\cdot\tfrac{w_k}{\kappa_k\sinh(w_k s)}\cdot\tfrac{g_k}{w_k}\tanh(w_k s/2)
= \tfrac{g_k}{2}\operatorname{sech}^2(w_k s/2)$ integrates exactly:
\begin{equation}
  \theta_{y;t}^{(-)}
  = \theta_{y;k+1}^{(-)}
    + \frac{g_k}{w_k}\tanh\!\Bigl(\frac{w_k(t_{k+1}-t)}{2}\Bigr)
  = \theta_{y;k+1}^{(-)} + \theta_{x;t}^{(-)},
  \qquad [\sigma_k=0,\,\beta_k=b_k I].
  \label{appA:eq:thetay_isotropic}
\end{equation}
No quadrature is needed; $\theta_y$ and $\theta_x$ accumulate
\emph{identical} increments across each interval.

\paragraph*{Zero-drift case $\sigma_k=0$, general $\beta_k\succ 0$.}
Replace the scalar $w_k$ by the matrix square root
$W_k\doteq(\kappa_k\beta_k)^{1/2}$ (symmetric positive definite).
Using matrix hyperbolic functions, the same derivation gives
\begin{equation}
  \theta_{x}^{(-)}(\tau)
  = W_k^{-1}\tanh\!\Bigl(\frac{W_k\tau}{2}\Bigr)g_k,
  \qquad
  \theta_{y;t}^{(-)}
  = \theta_{y;k+1}^{(-)} + W_k^{-1}\tanh\!\Bigl(\frac{W_k(t_{k+1}-t)}{2}\Bigr)g_k,
  \label{appA:eq:thetay_zero_drift}
\end{equation}
where $W_k^{-1}\tanh(W_k\tau/2)$ denotes the matrix function evaluated
via the eigendecomposition $\beta_k=Q_k D_k Q_k^\top$: component $i$
gets the scalar factor $\tanh(w_{k,i}\tau/2)/w_{k,i}$ with
$w_{k,i}=\sqrt{\kappa_k(D_k)_{ii}}$. Again no quadrature is required.

\paragraph*{Scalar-drift case $\sigma_k = c_k I$, diagonal $\beta_k =
\mathrm{diag}(b_{k,1},\ldots,b_{k,d})$.}
When $\sigma_k = c_k I$ (a scalar multiple of the identity), the
backward Hamiltonian $-H_k^{(-)}$ has the block form
\[
  -H_k^{(-)}
  = \begin{pmatrix} -c_k I & \kappa_k I \\ \beta_k & c_k I \end{pmatrix},
\]
which still decouples into $d$ independent $2\times 2$ systems.
Mode $i$ has eigenvalues $\pm w_{k,i}$ with
\begin{equation}
  w_{k,i} \doteq \sqrt{\kappa_k\,b_{k,i} + c_k^2}.
  \label{appA:eq:w_scalar_drift}
\end{equation}
The transition blocks (per mode) are
\begin{align}
  \Phi^{(-)}_{11,i}(\tau)
    &= \cosh(w_{k,i}\tau) - \tfrac{c_k}{w_{k,i}}\sinh(w_{k,i}\tau),
  \notag\\
  \Phi^{(-)}_{12,i}(\tau)
    &= \tfrac{\kappa_k}{w_{k,i}}\sinh(w_{k,i}\tau),
  \qquad
  \Phi^{(-)}_{22,i}(\tau)
    = \cosh(w_{k,i}\tau) + \tfrac{c_k}{w_{k,i}}\sinh(w_{k,i}\tau).
  \label{appA:eq:Phi_scalar_drift}
\end{align}
Because $c_k I$ commutes with every diagonal matrix, the mode
decoupling is preserved exactly as in the zero-drift case; the only
change is the frequency shift
$\sqrt{\kappa_k\,b_{k,i}} \to w_{k,i}$.
The $\theta_x$ recovery formula Eq.~(\ref{appA:eq:pwc_thetax_minus})
continues to hold with $X_k^{(-)}(\tau) =
\mathrm{diag}(\Phi^{(-)}_{11,i}) +
\mathrm{diag}(\Phi^{(-)}_{12,i})\,A_{k+1}^{(-)}$ and the augmented
drive integrals
\begin{equation}
  \widetilde\Gamma_{1,i}
    = g_i\!\left(\frac{\sinh(w_{k,i}\tau)}{w_{k,i}}
                  - c_k\frac{\cosh(w_{k,i}\tau)-1}{w_{k,i}^2}\right),
  \qquad
  \widetilde\Gamma_{2,i}
    = g_i\,\kappa_k\,\frac{\cosh(w_{k,i}\tau)-1}{w_{k,i}^2}.
  \label{appA:eq:Gtilde_scalar_drift}
\end{equation}
The $\theta_y$ integral closes exactly as in the zero-drift case.
At the terminal ($t_{k+1}$) boundary where $\theta_{x;k+1}^{(-)}=0$,
substituting $B^{(-)}(s)_i = w_{k,i}/(\kappa_k\sinh(w_{k,i}s))$ and
$\theta_x^{(-)}(s)_i = (g_i/w_{k,i})\tanh(w_{k,i}s/2)$ into the
integrand gives $\kappa_k\cdot(g_i/2\kappa_k)\operatorname{sech}^2(w_{k,i}s/2)
=(g_i/2)\operatorname{sech}^2(w_{k,i}s/2)$, which integrates exactly:
\begin{equation}
  \theta_{y;t}^{(-)}
  = \theta_{y;k+1}^{(-)}
    + \frac{g_k \mathbin{\odot} \tanh\!\bigl(\mathbf{w}_k(t_{k+1}-t)/2\bigr)}%
           {\mathbf{w}_k},
  \qquad [\sigma_k = c_k I,\;\beta_k\text{ diagonal}],
  \label{appA:eq:thetay_scalar_drift}
\end{equation}
where $\mathbf{w}_k = (w_{k,1},\ldots,w_{k,d})^\top$ with
$w_{k,i}=\sqrt{\kappa_k\,b_{k,i}+c_k^2}$, $g_k = \beta_k\nu_k$, and
$\odot$ denotes elementwise division/multiplication. The formula is
identical to Eq.~(\ref{appA:eq:thetay_zero_drift}) with the frequency shift
$\sqrt{\kappa_k\,b_{k,i}}\to w_{k,i}$. For general SPD $\beta_k$ with
scalar drift $\sigma_k=c_kI$, diagonalise $\beta_k = Q_k D_k Q_k^\top$
and apply the same formula mode-wise with
$w_{k,i}=\sqrt{\kappa_k\,(D_k)_{ii}+c_k^2}$.

\paragraph*{General $\sigma_k$, $\beta_k$.}
When $\sigma_k\neq 0$ (and not a scalar multiple of $I$ commuting with
$\beta_k$), the two equations for $\theta_x$ and $\theta_y$ no longer
decouple from $A_t^{(-)}$ in the same simple way, and the integral
Eq.~(\ref{appA:eq:pwc_thetay_minus}) does not reduce to a matrix-function
evaluation; Gauss--Legendre quadrature is used to evaluate the
integral in Eq.~(\ref{appA:eq:pwc_thetay_minus}).

\subsubsection{Forward branch: explicit update on an interval}

On $I_k$ define the forward-branch Hamiltonian matrix
\begin{equation}
  H_k^{(+)}
  \doteq
  \begin{pmatrix}
    \sigma_k & \kappa_k I\\
    \beta_k & -\sigma_k^\top
  \end{pmatrix},
  \qquad
  \Phi_k^{(+)}(\tau)\doteq \exp\!\bigl(H_k^{(+)}\,\tau\bigr)
  =
  \begin{pmatrix}
    \Phi^{(+)}_{11}(\tau) & \Phi^{(+)}_{12}(\tau)\\
    \Phi^{(+)}_{21}(\tau) & \Phi^{(+)}_{22}(\tau)
  \end{pmatrix}.
  \label{appA:eq:Hplus_def}
\end{equation}

\paragraph*{Initialization and induction across intervals.}
The forward recursion is initialized by the delta asymptotics at
$t\downarrow 0$,
Eq.~(\ref{appA:eq:bc_plus_ABC_clean}--\ref{appA:eq:bc_plus_theta_clean}).
In practice, one evaluates the coefficients on $t\ge \varepsilon$ and
starts the recursion at a time $t_\star=\varepsilon$ with large values
consistent with the asymptotic (e.g.\
$A_{t_\star}^{(+)}\approx (\kappa_0\varepsilon)^{-1}I$,
$B_{t_\star}^{(+)}\approx (\kappa_0\varepsilon)^{-1}I$,
$C_{t_\star}^{(+)}\approx (\kappa_0\varepsilon)^{-1}I$ and
$\theta_{x;t_\star}^{(+)}=\theta_{y;t_\star}^{(+)}=0$). Then enforce
continuity at each grid point and use the right-end values from $I_k$
as the left-end initial data for $I_{k+1}$.

\paragraph*{Quadratic coefficients (explicit).}
Fix initial values at $t_k$:
\[
A_k^{(+)} \doteq A_{t_k}^{(+)},\quad
B_k^{(+)} \doteq B_{t_k}^{(+)},\quad
C_k^{(+)} \doteq C_{t_k}^{(+)}.
\]
Define
\begin{align}
  X_k^{(+)}(\tau) &\doteq \Phi^{(+)}_{11}(\tau) + \Phi^{(+)}_{12}(\tau)\,A_k^{(+)},
  \\
  Y_k^{(+)}(\tau) &\doteq \Phi^{(+)}_{21}(\tau) + \Phi^{(+)}_{22}(\tau)\,A_k^{(+)}.
\end{align}
Then for $t=t_k+\tau$,
\begin{align}
  A_t^{(+)} &= Y_k^{(+)}(\tau)\,\bigl(X_k^{(+)}(\tau)\bigr)^{-1},
  \label{appA:eq:pwc_A_plus}
  \\
  B_t^{(+)} &= \bigl(X_k^{(+)}(\tau)\bigr)^{-\top}\,B_k^{(+)}.
  \label{appA:eq:pwc_B_plus}
\end{align}
Closed-form updates for $C_t^{(+)}$, $\theta_{x;t}^{(+)}$, and
$\theta_{y;t}^{(+)}$ are obtained analogously by using the same
augmentation pattern as in the backward branch:
\begin{itemize}
  \item for $C^{(+)}$: use an augmented matrix identical to
    Eq.~(\ref{appA:eq:Hminus_aug_C}) but with the forward Hamiltonian
    blocks Eq.~(\ref{appA:eq:Hplus_def};
  \item for $\theta_x^{(+)}$: use an augmented matrix identical to
    Eq.~(\ref{appA:eq:Hminus_aug_theta}) but with the forward Hamiltonian
    blocks and $g_k$ in the bottom-left block (the drive is $+g_k$,
    consistent with Eq.~(\ref{appA:eq:ode_thetax_plus_clean}); recovery
    via Eq.~(\ref{appA:eq:pwc_thetax_minus}) with $A_{k+1}^{(-)}\to A_k^{(+)}$,
    $\theta_{x;k+1}^{(-)}\to\theta_{x;k}^{(+)}$;
  \item for $\theta_y^{(+)}$: use the integral formula
    Eq.~(\ref{appA:eq:pwc_thetay_minus}).
\end{itemize}

\subsection{From \texorpdfstring{$\delta$}{delta}-source to Gaussian-Mixture: Explicit Expressions}
\label{appA:gmm_target_explicit}

In this subsection we assume a Gaussian-mixture target at $t=T$:
\begin{equation}
  p^{(\mathrm{tar})}(y)
  = \sum_{k=1}^K \pi_k\,\mathcal N(y; m_k,\Sigma_k),
  \qquad \pi_k>0,\ \sum_k\pi_k=1,
  \label{appA:eq:gmm_target}
\end{equation}
with $\Sigma_k\succ0$ and precision $P_k\doteq \Sigma_k^{-1}$.

In the case of the $\delta$-source,
Eqs.~Eq.~(\ref{appA:eq:p_star-via-G}--\ref{appA:eq:u_star-via-G}) become
\begin{align}\label{appA:eq:phi-psi-delta}
     \varphi_t(x) &= G_t^{(+)}(x\mid 0),\quad
  \psi_t(x)  \propto\ \int_{\R^d}
    \frac{p^{(\mathrm{tar})}(y)}{\varphi_T(y)}\, G_t^{(-)}(x\mid y)\,\dd y,
  \\ \label{appA:eq:p_star-via-G-delta}
  p_t^{(*)}(x) & \propto \varphi_t(x) \psi_t(x)
  \propto \left(
  \int_{\R^d} G_t^{(+)}(x\mid z)\,p^{(\mathrm{in})}(z)\,\dd z\right)
  \left(\int_{\R^d}
    \frac{p^{(\mathrm{tar})}(y) G_t^{(-)}(x\mid y)}{ G_T^{(+)}(y\mid 0)}\,\dd y\right),
  \\ \label{appA:eq:u_star-via-G-delta}
  u_t^*(x) &= \kappa_t \nabla_x \log\left(\int_{\R^d}
    \frac{p^{(\mathrm{tar})}(y) G_t^{(-)}(x\mid y)}{ G_T^{(+)}(y\mid 0)}\,\dd y\right).
\end{align}

The integrand appearing in the backward message has a useful probabilistic interpretation. For fixed $(t,x)$ define the normalized terminal \emph{look-up law}
\begin{equation}
  \ell_t(y\mid x)
  \doteq
  \frac{\displaystyle p^{(\mathrm{tar})}(y)\,G_t^{(-)}(x\mid y) / G_T^{(+)}(y\mid x_0)}
       {\displaystyle \int_{\R^d} p^{(\mathrm{tar})}(y')\,G_t^{(-)}(x\mid y') / G_T^{(+)}(y'\mid x_0)\,\dd y'} .
  \label{appA:eq:lookup_law}
\end{equation}
This is the bridge-induced distribution over terminal endpoints $y=X_T$ compatible with the intermediate observation $X_t=x$. Equivalently, it is a finite-time Doob-transform analogue of a posterior over the final endpoint: the numerator combines the target weight of $y$, the backward likelihood of observing $x$ at time $t$ when the endpoint is $y$, and the normalization by the uncontrolled forward reachability of $y$ from the source. The corresponding look-up expectation is
\begin{equation}
  \widehat x_T(t;x)
  \doteq
  \E_{Y\sim \ell_t(\cdot\mid x)}[Y]
  =
  \int_{\R^d} y\,\ell_t(y\mid x)\,\dd y .
  \label{appA:eq:lookup_expectation}
\end{equation}
It maps a current state $x$ to the terminal endpoint predicted by the bridge at time $t$. In multimodal targets, the associated component responsibilities quantify terminal-mode commitment and are therefore useful diagnostics of branch selection and dynamic phase-transition-like behavior.

Substitution of Eq.~(\ref{appA:eq:gmm_target}) into these formulas results,
after taking Gaussian integrals, in explicit algebraic expressions.

\subsubsection{Closed form for the backward message}

Fix any $t\in(0,T)$ (in coding: $t$ lies in some PWC interval where
the coefficient functions are already available explicitly).
We use the coefficient representation Eq.~(\ref{appA:eq:pm_ansatz_clean})
for $G_t^{(-)}(x\mid y)$ and for $G_T^{(+)}(y\mid x_0)$:
\begin{align}
  G_t^{(-)}(x\mid y)
  &=
  \exp\!\Big(
      -\tfrac12 x^\top A_t^{(-)} x
      + x^\top B_t^{(-)} y
      -\tfrac12 y^\top C_t^{(-)} y
      + (\theta_{x;t}^{(-)})^\top x
      + (\theta_{y;t}^{(-)})^\top y
      + \zeta_t^{(-)}
  \Big), \label{appA:eq:Gminus_here}\\
  G_T^{(+)}(y\mid x_0)
  &=
  \exp\!\Big(
      -\tfrac12 y^\top A_T^{(+)} y
      + y^\top B_T^{(+)} x_0
      + (\theta_{x;T}^{(+)})^\top y
      + \zeta_T^{(+)}(x_0)
  \Big). \label{appA:eq:Gplus1_here}
\end{align}
(Only the $y$-dependence of $G_T^{(+)}$ matters; the term
$\zeta_T^{(+)}(x_0)$ is $y$-independent.)

Define, for each mixture component $k$, the symmetric matrix
\begin{equation}
  S_{k,t}
  \doteq C_t^{(-)} + P_k - A_T^{(+)},
  \qquad S_{k,t}=S_{k,t}^\top,
  \label{appA:eq:Skt_def}
\end{equation}
and the vector
\begin{equation}
  q_{k,t}
  \doteq
  \theta_{y;t}^{(-)} + P_k m_k - B_T^{(+)}x_0 - \theta_{x;T}^{(+)}.
  \label{appA:eq:qkt_def}
\end{equation}
Define
\begin{equation}
  \psi_t(x)\ \propto\ \sum_{k=1}^K \psi_{k,t}(x).
  \label{appA:eq:psi_sum}
\end{equation}
Then the $k$th contribution to Eq.~(\ref{appA:eq:psi_sum}) is (up to a
$t$-only constant)
\begin{equation}
  \psi_{k,t}(x)
  \doteq
  \omega_{k,t}\,
  \exp\!\Big(
      -\tfrac12 x^\top \Lambda_{k,t} x
      + \lambda_{k,t}^\top x
  \Big),
  \label{appA:eq:psi_k_form}
\end{equation}
where
\begin{align}
  \Lambda_{k,t}
  &\doteq
  A_t^{(-)} - B_t^{(-)}\,S_{k,t}^{-1}\,(B_t^{(-)})^\top,
  \label{appA:eq:Lambda_kt}
  \\
  \lambda_{k,t}
  &\doteq
  \theta_{x;t}^{(-)} + B_t^{(-)}\,S_{k,t}^{-1}\,q_{k,t},
  \label{appA:eq:lambda_kt}
\end{align}
and the (explicit) scalar weight factor can be taken as
\begin{equation}
  \omega_{k,t}
  \doteq
  \pi_k\,|\Sigma_k|^{-1/2}\,|S_{k,t}|^{-1/2}\,
  \exp\!\Big(
      \zeta_t^{(-)} - \zeta_T^{(+)}(x_0)
      -\tfrac12 m_k^\top P_k m_k
      + \tfrac12 q_{k,t}^\top S_{k,t}^{-1} q_{k,t}
  \Big).
  \label{appA:eq:omega_kt}
\end{equation}

\subsubsection{Optimal control}

Since $u_t^*(x)=\kappa_t\nabla\log\psi_t(x)$ and each $\psi_{k,t}$ is
exponential-quadratic in $x$,
\begin{equation}
  \nabla\log \psi_{k,t}(x) = -\Lambda_{k,t}x + \lambda_{k,t}.
\end{equation}
Introduce the time-$t$ responsibilities
\begin{equation}
  \rho_{k,t}(x)
  \doteq
  \frac{\psi_{k,t}(x)}{\sum_{j=1}^K \psi_{j,t}(x)}\in(0,1),
  \qquad \sum_{k=1}^K \rho_{k,t}(x)=1.
  \label{appA:eq:rho_def}
\end{equation}
Then the optimal control is the explicit convex combination
\begin{equation}
  u_t^*(x)
  = \kappa_t\sum_{k=1}^K \rho_{k,t}(x)\,\bigl(-\Lambda_{k,t}x + \lambda_{k,t}\bigr).
  \label{appA:eq:u_star_gmm}
\end{equation}

\subsubsection{Closed form for the terminal look-up map}
\label{appA:lookup_law}

The same Gaussian integration gives an explicit expression for the look-up law Eq.~(\ref{appA:eq:lookup_law}). Conditional on terminal mixture component $k$, the normalized density over $y$ is Gaussian with precision $S_{k,t}$ and mean
\begin{equation}
  \widehat y_{k,t}(x)
  \doteq
  S_{k,t}^{-1}\bigl((B_t^{(-)})^\top x + q_{k,t}\bigr).
  \label{appA:eq:lookup_component_mean}
\end{equation}
The component weights are precisely the responsibilities $\rho_{k,t}(x)$ in Eq.~(\ref{appA:eq:rho_def}). Hence
\begin{equation}
  \ell_t(y\mid x)
  =
  \sum_{k=1}^K \rho_{k,t}(x)\,
  \mathcal N\!\bigl(y;\widehat y_{k,t}(x),S_{k,t}^{-1}\bigr),
  \label{appA:eq:lookup_gmm}
\end{equation}
up to the standard Gaussian normalizing convention already absorbed in $\omega_{k,t}$. The corresponding expectation is the closed-form terminal look-up map
\begin{equation}
  \widehat x_T(t;x)
  =
  \sum_{k=1}^K \rho_{k,t}(x)\,\widehat y_{k,t}(x)
  =
  \sum_{k=1}^K \rho_{k,t}(x)\,
  S_{k,t}^{-1}\bigl((B_t^{(-)})^\top x + q_{k,t}\bigr).
  \label{appA:eq:lookup_expectation_gmm}
\end{equation}
Thus the same quantities used to evaluate the score also provide an inference-time terminal prediction. A sharp temporal transition in the responsibilities $\rho_{k,t}(x)$ or in the population average of $\widehat x_T(t;X_t)$ signals rapid commitment of the bridge ensemble to terminal modes.

\subsubsection{Optimal density}

With deterministic start, Eq.~(\ref{appA:eq:p_star-via-G-delta}) and
Eq.~(\ref{appA:eq:psi_sum}) imply
\begin{equation}
  p_t^*(x) \ \propto\ G_t^{(+)}(x\mid x_0)\,\sum_{k=1}^K \psi_{k,t}(x).
  \label{appA:eq:p_star_prop}
\end{equation}
Using the Gaussian-exponential form for $G_t^{(+)}(x\mid x_0)$:
\begin{equation}
  G_t^{(+)}(x\mid x_0)
  =
  \exp\!\Big(
      -\tfrac12 x^\top A_t^{(+)} x
      + x^\top B_t^{(+)} x_0
      + (\theta_{x;t}^{(+)})^\top x
      + \zeta_t^{(+)}(x_0)
  \Big),
  \label{appA:eq:Gplus_here}
\end{equation}
each product $G_t^{(+)}(x\mid x_0)\psi_{k,t}(x)$ is again
exponential-quadratic in $x$ and hence a Gaussian (up to normalization).
Define the component precision and mean:
\begin{align}
  \Pi_{k,t}
  &\doteq
  A_t^{(+)} + \Lambda_{k,t},
  \qquad \Pi_{k,t}=\Pi_{k,t}^\top,
  \label{appA:eq:Pi_kt}
  \\
  \mu_{k,t}
  &\doteq
  \Pi_{k,t}^{-1}\Big(B_t^{(+)}x_0 + \theta_{x;t}^{(+)} + \lambda_{k,t}\Big).
  \label{appA:eq:mu_kt}
\end{align}
Then
\begin{equation}
  p_t^*(x)
  =
  \sum_{k=1}^K \bar\pi_{k,t}\,\mathcal N\!\bigl(x;\mu_{k,t},\Pi_{k,t}^{-1}\bigr),
  \label{appA:eq:p_star_gmm}
\end{equation}
where the (time-$t$) mixture weights are
\begin{equation}
  \bar\pi_{k,t}
  \doteq
  \frac{\widetilde\omega_{k,t}}{\sum_{j=1}^K \widetilde\omega_{j,t}},
  \qquad
  \widetilde\omega_{k,t}
  \doteq
  \omega_{k,t}\,|\Pi_{k,t}|^{-1/2}
  \exp\!\Big(\tfrac12 \mu_{k,t}^\top \Pi_{k,t}\mu_{k,t}\Big),
  \label{appA:eq:pi_bar}
\end{equation}
and $\omega_{k,t}$ is given in Eq.~(\ref{appA:eq:omega_kt}). (Any common
$k$-independent factor cancels in Eq.~(\ref{appA:eq:pi_bar}).)

\paragraph*{Summary.}
Given PWC closed-form coefficient functions on each interval, compute
$S_{k,t},q_{k,t},\Lambda_{k,t},\lambda_{k,t}$ via
Eq.~(\ref{appA:eq:Skt_def}--\ref{appA:eq:lambda_kt}), then evaluate
$u_t^*(x)$ by Eq.~(\ref{appA:eq:u_star_gmm}) and $p_t^*(x)$ by
Eq.~(\ref{appA:eq:p_star_gmm}--\ref{appA:eq:pi_bar}).

\subsubsection{Plug-and-play recipe}

This subsection summarizes the construction and evaluation pipeline
in one place.

\paragraph*{Input.}
A PWC grid and parameters
$\{(\sigma_k,\beta_k,\nu_k,\kappa_k)\}_{k=0}^{K-1}$, deterministic
start $x_0$, and a Gaussian-mixture target
$\{(\pi_k,m_k,\Sigma_k)\}_{k=1}^K$.

\paragraph*{Coefficient construction on $t\in[\varepsilon,T-\varepsilon]$.}
Choose a small $\varepsilon>0$ and work away from endpoint singularities.
\begin{itemize}
\item Forward branch: initialize near $t=0$ consistently with the
delta asymptotic
Eq.~(\ref{appA:eq:bc_plus_ABC_clean}--\ref{appA:eq:bc_plus_theta_clean})
(e.g.\ $A^{(+)},B^{(+)},C^{(+)}\approx (\kappa_0\varepsilon)^{-1}I$ and
$\theta_x^{(+)},\theta_y^{(+)}=0$ at $t=\varepsilon$), then propagate
forward interval-by-interval using \S\ref{appA:pwc_closed_form}. Cache
the terminal $y$-dependent coefficients
$A_{T-\varepsilon}^{(+)}$, $B_{T-\varepsilon}^{(+)}$,
$\theta_{x;T-\varepsilon}^{(+)}$.

\item Backward branch: initialize near $t=T$ consistently with the
delta asymptotic
Eq.~(\ref{appA:eq:bc_minus_ABC_clean}--\ref{appA:eq:bc_minus_theta_clean})
(e.g.\ $A^{(-)},B^{(-)},C^{(-)}\approx (\kappa_K\varepsilon)^{-1}I$ and
$\theta_x^{(-)},\theta_y^{(-)}=0$ at $t=T-\varepsilon$), then propagate
backward interval-by-interval using \S\ref{appA:pwc_closed_form}.

\item Enforce continuity of all coefficients across grid points.
\end{itemize}

\paragraph*{Evaluation of $\psi_t$, $u_t^*$ and $p_t^*$.}
For any query time $t\in[\varepsilon,T-\varepsilon]$:
\begin{itemize}
\item Use backward coefficients at $t$ and substitute
$(A_T^{(+)},B_T^{(+)},\theta_{x;T}^{(+)})\leftarrow
 (A_{T-\varepsilon}^{(+)},B_{T-\varepsilon}^{(+)},\theta_{x;T-\varepsilon}^{(+)})$
in Eq.~(\ref{appA:eq:Skt_def}--\ref{appA:eq:qkt_def}).

\item Compute $\psi_t(x)$ via Eq.~(\ref{appA:eq:psi_sum}), then $u_t^*(x)$
via Eq.~(\ref{appA:eq:u_star_gmm}).

\item Optionally compute $p_t^*(x)$ via
Eq.~(\ref{appA:eq:p_star_gmm}--\ref{appA:eq:pi_bar}).
\end{itemize}

\subsubsection{Algorithmic box}
\label{appA:algo_box}

\begin{center}
\fbox{
\begin{minipage}{0.96\linewidth}
\small
\textbf{Algorithm: PWC PID with a Gaussian-mixture target}\\[2pt]
\textbf{Inputs:}
grid $0=t_0<\cdots<t_K=T$; PWC parameters
$\{(\sigma_k,\beta_k,\nu_k,\kappa_k)\}_{k=0}^{K-1}$;
start $x_0\in\R^d$; target $\{(\pi_k,m_k,\Sigma_k)\}_{k=1}^K$;
query band $t\in[\varepsilon,T-\varepsilon]$.\\
\textbf{Outputs:}
for any $t\in[\varepsilon,T-\varepsilon]$: evaluators for
$\psi_t(x)$, $u_t^*(x)$, and $p_t^*(x)$.

\vspace{4pt}
\textbf{(A) Forward sweep and cache.}
\begin{enumerate}
\item Initialize at $t=\varepsilon$ with large values consistent with
Eq.~(\ref{appA:eq:bc_plus_ABC_clean}--\ref{appA:eq:bc_plus_theta_clean}).

\item Propagate forward coefficients on each interval using
\S\ref{appA:pwc_closed_form}, enforcing continuity.

\item Cache $A_{T-\varepsilon}^{(+)}$, $B_{T-\varepsilon}^{(+)}$,
$\theta_{x;T-\varepsilon}^{(+)}$.
\end{enumerate}

\vspace{4pt}
\textbf{(B) Backward sweep.}
\begin{enumerate}
\item Initialize at $t=T-\varepsilon$ with large values consistent with
Eq.~(\ref{appA:eq:bc_minus_ABC_clean}--\ref{appA:eq:bc_minus_theta_clean}).

\item Propagate backward coefficients on each interval using
\S\ref{appA:pwc_closed_form}, enforcing continuity.
\end{enumerate}

\vspace{4pt}
\textbf{(C) Evaluate at time $t$.}
\begin{enumerate}
\item Precompute $P_k=\Sigma_k^{-1}$.

\item Form $S_{k,t}=C_t^{(-)}+P_k-A_{T-\varepsilon}^{(+)}$ and
$q_{k,t}=\theta_{y;t}^{(-)}+P_k m_k - B_{T-\varepsilon}^{(+)}x_0-\theta_{x;T-\varepsilon}^{(+)}$.

\item Compute $\Lambda_{k,t}$ and $\lambda_{k,t}$ via
Eq.~(\ref{appA:eq:Lambda_kt}--\ref{appA:eq:lambda_kt}) (use linear solves).

\item Compute responsibilities by log-sum-exp from
$\log\psi_{k,t}(x)$ in Eq.~(\ref{appA:eq:psi_k_form}).

\item Output $u_t^*(x)=\kappa_t\sum_k \rho_{k,t}(x)(-\Lambda_{k,t}x+\lambda_{k,t})$;
optionally compute $p_t^*(x)$.
\end{enumerate}
\end{minipage}
}
\end{center}

\subsection{Finite empirical sources: sample-conditioned deterministic-start reduction}
\label{appA:gmm_initial}

\begin{revision}
In \S\ref{appA:gmm_target_explicit} we assumed a deterministic start. For a genuinely non-degenerate source density $p^{(\mathrm{in})}$, substituting both endpoint marginals into the ratio formulas does \emph{not} in general reduce to the finite Gaussian algebra of the deterministic-start case. Enforcing both continuous endpoint marginals requires a second Schrödinger scaling (equivalently, a Sinkhorn/iterative-proportional-fitting step in the discretized problem). We therefore do not claim a closed-form Gaussian-mixture-source theorem.

The E2 experiment instead draws $B$ points $z^{(n)}\sim p^{(\mathrm{in})}$ once and replaces the source by the empirical measure $B^{-1}\sum_n\delta(\cdot-z^{(n)})$. For each atom, all deterministic-start formulas of \S\ref{appA:gmm_target_explicit} apply exactly. The reported finite-source marginal is the equal-weight average of these normalized conditional bridge marginals. This is a computational reduction and an empirical approximation of the source law, not an analytic solution of the continuous two-marginal problem.
\end{revision}

\subsubsection{Coordinate-shift reduction}
\label{appA:coord_shift}

\begin{revision}The shift shortcut below is used only for E2, where $\sigma_t\equiv0$. If $f_t(x)=\sigma_t x$ with nonzero $\sigma_t$, the change $x=\tilde x+z$ generates an affine term $\sigma_t z$; the formulas below would then require an affine-drift extension. We therefore restrict this subsection's coordinate-shift identities to $\sigma_t=0$.\end{revision}

The central observation is that a coordinate shift $\tilde x = x - z$
with fixed $z\in\R^d$ transforms the zero-$\sigma$ LQ problem
Eq.~(\ref{appA:eq:linear_drift_quadratic_potential_user}) into an identical
LQ problem with modified parameters:
\begin{equation}
  \tilde\sigma_k=0,\qquad
  \tilde\beta_k=\beta_k,\qquad
  \tilde\kappa_k=\kappa_k,\qquad
  \tilde\nu_k = \nu_k - z,\qquad
  \tilde m_j = m_j - z,
  \label{appA:eq:shift_params}
\end{equation}
while all other quantities (target weights $\pi_k$, target covariances
$\Sigma_k$) are unchanged. In shifted coordinates the initial condition
is $\tilde x_0 = 0$ (deterministic), and the standard $\delta$-function
boundary condition applies at $t=0$.

\paragraph*{Effect on coefficients.}
Since $\sigma$, $\beta$, and $\kappa$ are unchanged, the
\emph{quadratic} Green-function coefficients $A^{(\pm)}_t$,
$B^{(\pm)}_t$, $C^{(\pm)}_t$ are identical for the shifted and
unshifted problems. Only the \emph{linear} coefficients
$\theta_{x;t}^{(\pm)}$, $\theta_{y;t}^{(\pm)}$ change, because they are
driven by $g_k = \beta_k\nu_k$: replacing $\nu_k\to \nu_k-z$ changes
$g_k\to g_k - \beta_k z$. Since the $\theta$ ODEs
Eq.~(\ref{appA:eq:ode_thetax_minus_clean}--\ref{appA:eq:ode_thetay_minus_clean})
(backward) and
Eq.~(\ref{appA:eq:ode_thetax_plus_clean}--\ref{appA:eq:ode_thetay_plus_clean})
(forward) are \emph{linear} in $g$, the shift is linear in $z$:
\begin{equation}
  \theta_{x;t}^{(\pm)}(z)
  = \theta_{x;t}^{(\pm)}\big|_{\nu}
    - \Lambda_{x;t}^{(\pm)}\,z,
  \qquad
  \theta_{y;t}^{(\pm)}(z)
  = \theta_{y;t}^{(\pm)}\big|_{\nu}
    - \Lambda_{y;t}^{(\pm)}\,z,
  \label{appA:eq:theta_shift_linear}
\end{equation}
where $\Lambda_{x;t}^{(\pm)},\Lambda_{y;t}^{(\pm)}\in\R^{d\times d}$ are
the \emph{shift-propagator matrices} defined below, and the subscript
``$|_{\nu}$'' denotes the base solution with the original $\nu_k$.

\subsubsection{Shift-propagator matrices}
\label{appA:shift_prop}

The shift-propagator matrices are defined as the $\theta$ responses to
unit coordinate drives. For each $j=1,\ldots,d$, consider the ``unit
protocol'' with $\nu_k = e_j$ (the $j$th standard basis vector) on
every interval, keeping $\sigma_k$, $\beta_k$, $\kappa_k$ fixed.
Denote the resulting backward and forward $\theta$ coefficients by
$\theta_{x;t}^{(-)}[e_j]$, $\theta_{y;t}^{(-)}[e_j]$,
$\theta_{x;t}^{(+)}[e_j]$ respectively. Then the shift-propagator
matrices are assembled column-by-column:
\begin{equation}
  \bigl(\Lambda_{x;t}^{(-)}\bigr)_{:,j}
  = \theta_{x;t}^{(-)}[e_j],
  \qquad
  \bigl(\Lambda_{y;t}^{(-)}\bigr)_{:,j}
  = \theta_{y;t}^{(-)}[e_j],
  \qquad
  \bigl(\Lambda_{x;T}^{(+)}\bigr)_{:,j}
  = \theta_{x;T}^{(+)}[e_j].
  \label{appA:eq:shift_prop_def}
\end{equation}
The $d$ unit sweeps share the same quadratic coefficients
$A^{(\pm)}$, $B^{(\pm)}$, $C^{(\pm)}$ as the base problem (since these
are $\nu$-independent), so only the linear-coefficient propagation
needs to be repeated---$d$ times for the backward branch and $d$ times
for the forward branch. This is a one-time $O(d)$ precomputation (per
sweep direction) on top of the base sweeps.

\paragraph*{Verification.}
The shift formula Eq.~(\ref{appA:eq:theta_shift_linear}) can be verified by
direct comparison: for any fixed $z$, build the full shifted protocol
Eq.~(\ref{appA:eq:shift_params}) and run a complete sweep. The resulting
$\theta_{x;t}^{(\pm)}$, $\theta_{y;t}^{(\pm)}$ must agree with the base
values minus the shift-propagator correction $\Lambda^{(\pm)}\,z$ to
machine precision. This is checked in our implementation for both
$z=0$ (identity) and nonzero $z$.

\subsubsection{Per-source atom $z$: conditional simulation}
\label{appA:per_particle_z}

Each source atom $z^{(n)}$ is drawn once from $p^{(\mathrm{in})}$
and runs in shifted coordinates $\tilde x^{(n)}=x^{(n)}-z^{(n)}$:
\begin{equation}
  \dd\tilde x^{(n)}_t
  = \tilde u^*_t\!\bigl(\tilde x^{(n)}_t;\,z^{(n)}\bigr)\dd t
    + \sqrt{\kappa_t}\,\dd W^{(n)}_t,
  \qquad
  \tilde x_0^{(n)} = 0.
  \label{appA:eq:sde_shifted}
\end{equation}
The control field $\tilde u^*_t(\tilde x;\,z)$ is evaluated using the
base quadratic quantities (shared across all particles) and
$z$-corrected linear quantities.

\paragraph*{Efficient control evaluation.}
At query time $t$, the base backward coefficients $A_t^{(-)}$,
$B_t^{(-)}$, $C_t^{(-)}$ and the terminal forward coefficient
$A_T^{(+)}$ are computed once. The per-component quantities from
\S\ref{appA:gmm_target_explicit} split into $z$-independent and
$z$-dependent parts:

\medskip\noindent
\emph{$z$-independent} (computed once):
\begin{equation}
  S_{k,t} = C_t^{(-)} + P_k - A_T^{(+)},
  \qquad
  \Lambda_{k,t} = A_t^{(-)} - B_t^{(-)}\,S_{k,t}^{-1}\,(B_t^{(-)})^\top.
\end{equation}
\emph{$z$-dependent} (per particle):
\begin{align}
  \tilde m_k(z)   &= m_k - z,
  \label{appA:eq:shifted_mk}
  \\
  \tilde q_{k,t}(z) &=
    \bigl(\theta_{y;t}^{(-)}\big|_\nu - \Lambda_{y;t}^{(-)}\,z\bigr)
    + P_k\,\tilde m_k(z)
    - \bigl(\theta_{x;T}^{(+)}\big|_\nu - \Lambda_{x;T}^{(+)}\,z\bigr),
  \label{appA:eq:shifted_qk}
  \\
  \tilde\lambda_{k,t}(z) &=
    \bigl(\theta_{x;t}^{(-)}\big|_\nu - \Lambda_{x;t}^{(-)}\,z\bigr)
    + B_t^{(-)}\,S_{k,t}^{-1}\,\tilde q_{k,t}(z),
  \label{appA:eq:shifted_lambdak}
  \\
  \tilde C_k(z) &=
    -\tfrac12\log|S_{k,t}| + \tfrac12\log|P_k|
    - \tfrac12 \tilde m_k(z)^\top P_k\,\tilde m_k(z)
    + \tfrac12 \tilde q_{k,t}(z)^\top S_{k,t}^{-1}\,\tilde q_{k,t}(z).
  \label{appA:eq:shifted_Ck}
\end{align}
The control follows from the standard formula
Eq.~(\ref{appA:eq:u_star_gmm}) with these shifted quantities:
\begin{equation}
  \tilde u_t^*(\tilde x;\,z)
  = \kappa_t \sum_{k=1}^K
    \tilde\rho_{k,t}(\tilde x;\,z)\,
    \bigl(-\Lambda_{k,t}\tilde x + \tilde\lambda_{k,t}(z)\bigr),
  \label{appA:eq:u_star_shifted}
\end{equation}
where $\tilde\rho_{k,t}$ are the responsibilities computed from the
$z$-corrected log-weights:
\begin{equation}
  \log \tilde w_k(\tilde x;\,z)
  = \log\pi_k + \tilde C_k(z)
    - \tfrac12 \tilde x^\top\Lambda_{k,t}\tilde x
    + \tilde\lambda_{k,t}(z)^\top \tilde x.
  \label{appA:eq:shifted_log_wk}
\end{equation}
The physical trajectory is recovered as $x^{(n)}_t = \tilde x^{(n)}_t + z^{(n)}$.

\subsubsection{Computational cost and algorithmic summary}
\label{appA:gm_gm_algo}

\paragraph*{Precomputation.} On top of the standard backward/forward sweeps (steps A--B in the algorithm of \S\ref{appA:algo_box}), one runs $d$ additional backward sweeps and $d$ additional forward sweeps with the unit protocols $\nu=e_j$, $j=1,\ldots,d$. Since the quadratic coefficients $A^{(\pm)}$, $B^{(\pm)}$, $C^{(\pm)}$ are identical to the base sweep, these extra sweeps only propagate the linear coefficients---the per-sweep cost is comparable to the base sweep. Total additional precomputation: $O(d)$ sweeps.

\paragraph*{Per-timestep evaluation.} At each Euler--Maruyama step, the $z$-independent quantities ($S_k$, $\Lambda_k$, $B^{(-)}S_k^{-1}$) are computed once in $O(K_{\mathrm{tar}}\cdot d^3)$ time, where $K_{\mathrm{tar}}$ is the number of target components. The shift-propagator matrices $\Lambda_{x;t}^{(-)}$, $\Lambda_{y;t}^{(-)}$ are evaluated by $d$ calls to the coefficient interpolator. For each particle, the $z$-correction involves $O(K_{\mathrm{tar}}\cdot d^2)$ operations (matrix--vector products and inner products). The total per-step cost is $O(K_{\mathrm{tar}}\cdot d^3 + B\cdot K_{\mathrm{tar}}\cdot d^2)$, where $B$ is the batch size---the same scaling as the deterministic-start algorithm, with an additional $O(d^2)$ per-particle overhead from the shift correction.

\begin{center}
\fbox{
\begin{minipage}{0.96\linewidth}
\small
\textbf{Algorithm: finite empirical source via deterministic-start conditionals}\\[2pt]
\textbf{Inputs:} PWC protocol $\{(\sigma_k,\beta_k,\nu_k,\kappa_k)\}$; initial GMM $\{(w_i^{(0)}, m_i^{(0)}, \Sigma_i^{(0)})\}_{i=1}^{M_0}$; target GMM $\{(\pi_k, m_k, \Sigma_k)\}_{k=1}^{K_{\mathrm{tar}}}$; batch size $B$; step count $N$.\\
\textbf{Outputs:}
terminal samples $\{x_T^{(n)}\}_{n=1}^B$ approximating $p^{(\mathrm{tar})}$.

\vspace{4pt}
\textbf{(A) Base sweeps.}
Run the standard backward and forward sweeps (\S\ref{appA:algo_box}, A--B) with $x_0=0$ and the original protocol $\{\nu_k\}$. Cache the backward states $\{A^{(-)}, B^{(-)}, C^{(-)}, \theta_x^{(-)}, \theta_y^{(-)}\}$ at breakpoints, and the terminal forward coefficients $A_T^{(+)}$, $\theta_{x;T}^{(+)}$.

\vspace{4pt}
\textbf{(B) Shift-propagator sweeps.} For $j=1,\ldots,d$: run backward and forward sweeps with $\nu_k = e_j$ on all intervals, storing $\theta_{x;t}^{(\pm)}[e_j]$ and $\theta_{y;t}^{(-)}[e_j]$ at breakpoints. Assemble the $(d\times d)$ matrices $\Lambda_{x;t}^{(-)}$, $\Lambda_{y;t}^{(-)}$, $\Lambda_{x;T}^{(+)}$ via Eq.~(\ref{appA:eq:shift_prop_def}).

\vspace{4pt}
\textbf{(C) Sample and simulate.}
\begin{enumerate}
\item For each particle $n=1,\ldots,B$: draw $z^{(n)}\sim p^{(\mathrm{in})}$
and set $\tilde x^{(n)}_0=0$.
\item For each timestep $i=0,\ldots,N-1$:
  \begin{enumerate}
  \item[(a)] Evaluate the base backward coefficients and shift-propagator
    matrices at $t_i$.
  \item[(b)] Compute $z$-independent quantities: $S_k$, $\Lambda_k$,
    $B^{(-)}S_k^{-1}$.
  \item[(c)] For each particle: apply $z$-corrections
    Eq.~(\ref{appA:eq:shifted_qk}--\ref{appA:eq:shifted_Ck}) and evaluate
    $\tilde u_t^*(\tilde x^{(n)}_t;\,z^{(n)})$ via
    Eq.~(\ref{appA:eq:u_star_shifted}).
  \item[(d)] Euler--Maruyama:
    $\tilde x^{(n)}_{t_{i+1}} = \tilde x^{(n)}_{t_i} + \tilde u^*\,\Delta t
      + \sqrt{\kappa\,\Delta t}\,\xi^{(n)}$.
  \end{enumerate}
\item Return physical trajectories $x^{(n)}_t = \tilde x^{(n)}_t + z^{(n)}$.
\end{enumerate}
\end{minipage}
}
\end{center}

\begin{revision}
\paragraph{Remark (conditional exactness vs. source approximation).}
For $\sigma_t=0$, each problem conditioned on a fixed $z^{(n)}$ is exactly a deterministic-start LQ-GM-PID bridge in continuous time. The finite mixture over $n$ is therefore analytic conditional on the empirical source, and each conditional bridge has the prescribed target law. The approximation is the replacement of the continuous source distribution by the empirical measure. A true bridge that enforces the continuous source and target marginals simultaneously is a different two-sided scaling problem and is not claimed to be closed form here.
\end{revision}

\section{Full E1 protocol-learning study}
\label{app:caseB}

This appendix gives the complete specification of the E1 demonstration
of \S\ref{sec:E1}: corridor geometry and target, optimization loss
structure, optimizer configuration, the loss-history curve, the
optimized protocol parameters. \begin{revision}\end{revision} All quantities are reproducible from a
fixed seed via the notebook \path{experiments/exp_E1_corridor.ipynb};
the configuration values listed below are the literal entries in that
notebook's first code cell.

\subsection{Geometry and target}

\paragraph{Corridor midline.}
\begin{revision}The corridor is a curved single arch. Define
\[
 h(s)=A\,[\tanh(\kappa(s-0.30))-\tanh(\kappa(s-0.70))],\qquad A=0.7,\quad \kappa=6,
\]
then detrend the vertical coordinate exactly as in the released code,
\[
 y(s)=h(s)-h(0)-s\,[h(1)-h(0)],\qquad m(s)=(3s,y(s)).
\]
This gives $m(0)=(0,0)$ and $m(1)=(3,0)$ and produces the smooth arch visible in the figures. The resulting centerline is the single arch visible in the figures.\end{revision}

\paragraph{Target.}
The terminal mixture is bimodal at the corridor exit:
\[
  p^{(\mathrm{tar})}(y)
  \;=\;
  \tfrac12\,\mathcal N\bigl(y;\,(3, +0.5),\,\mathrm{diag}(0.06, 0.05)\bigr)
  +
  \tfrac12\,\mathcal N\bigl(y;\,(3, -0.5),\,\mathrm{diag}(0.06, 0.05)\bigr).
\]

\paragraph{Source.}
Deterministic start at $x_0 = (0, 0)$.

\subsection{Trainable protocol}

The protocol uses $K = 10$ uniform PWC intervals on $[0,1]$. Each
interval $k$ has midpoint $s_k = (k - \tfrac12)/K$ and is
parameterized by two scalar variables:
\begin{itemize}[leftmargin=1.5em]
  \item Transverse guide offset $\rho_k \in \R$ (no constraint), giving
  the guide centerline $\nu_k = m(s_k) + \rho_k\, n(s_k)$, where
  $n(s_k)$ is the corridor unit normal.
  \item Unconstrained $c_k \in \R$, giving the transverse stiffness via
  the sigmoid reparameterization
  \[
    \beta_k^{(\perp)}
    \;=\;
    \beta_{\min}^{(\perp)} + \bigl(\beta_{\max}^{(\perp)} - \beta_{\min}^{(\perp)}\bigr)\,
    \mathsf{sigmoid}(c_k),
    \qquad
    \bigl[\beta_{\min}^{(\perp)},\beta_{\max}^{(\perp)}\bigr] = [2,\,60].
  \]
\end{itemize}
The longitudinal stiffness $\beta^{(\parallel)} = 0.2$ is fixed.
The full shaping matrix on interval $k$ is
\[
  \beta_k
  \;=\;
  Q_k\,\mathrm{diag}\bigl(\beta^{(\parallel)},\,\beta_k^{(\perp)}\bigr)\,Q_k^\top,
  \qquad Q_k = [\,t(s_k)\,|\,n(s_k)\,],
\]
i.e.\ a corridor-aligned anisotropic stiffness with the longitudinal
direction permissive and the transverse direction tight (and learnable).
The state-dependent drift is $\sigma_t \equiv 0$, the noise schedule is
$\kappa_t \equiv 1$, and neither is trained.

\paragraph{Total parameter count.}
$2K = 20$ scalar parameters. The trainable state has $\dim(\rho) =
\dim(c) = 10$.

\subsection{Loss}

The total loss is
\[
  \mathcal L(\Gamma)
  \;=\;
  \lambda_{\mathrm{corr}}\,\mathcal L_{\mathrm{corr}}
  \;+\;
  \lambda_\rho\,\mathcal L_\rho
  \;+\;
  \lambda_\beta\,\mathcal L_\beta,
\]
with weights
$(\lambda_{\mathrm{corr}}, \lambda_\rho, \lambda_\beta) = (10,\,0.10,\,0.05)$.
The corridor-alignment term is Eq.~(\ref{eq:E1_loss}), with corridor
kernel widths $(\omega_\parallel, \omega_\perp) = (0.8,\,0.2)$ and the
active window $\mathcal K = \{k : s_k \le t_* = 0.80\}$. The
regularizers are
\begin{align*}
  \mathcal L_\rho
  ={}& \tfrac1{K-1}\sum_{k=1}^{K-1}(\rho_{k+1}-\rho_k)^2
  + 0.5\!\cdot\!\tfrac1{K-2}\sum_{k=2}^{K-1}(\rho_{k+1}-2\rho_k+\rho_{k-1})^2 \\
  &+ 0.3\!\cdot\!\tfrac1K\sum_{k=1}^K
  \mathsf{softplus}\!\bigl(15(|\rho_k|-0.55)\bigr)/15 .
\end{align*}
\[
  \mathcal L_\beta
  \;=\;
  \tfrac1{K-1}\sum_{k=1}^{K-1}(c_{k+1}-c_k)^2
  + \tfrac1K\sum_{k=1}^K (c_k-c_k^{(0)})^2,
\]
where $c_k^{(0)} = \mathsf{sigmoid}^{-1}\bigl((15-2)/(60-2)\bigr)$ is
the warm-start value (the inverse-sigmoid of $\beta_{\mathrm{init}}^{(\perp)} =
15$ inside the $[2,60]$ band). $\mathcal L_\rho$ penalizes large variation between neighboring PWC
guide values with a soft barrier at $|\rho| > 0.55$;
$\mathcal L_\beta$ keeps the transverse stiffness near the warm-start
unless the corridor-alignment objective explicitly benefits from
deviating.

\subsection{Optimizer}

Plain gradient descent uses learning rate $\eta=3\times10^{-2}$ for $300$ iterations, with no momentum. Each iteration performs the analytic Riccati/marginal calculation followed by reverse-mode differentiation; there is no SDE simulation inside the optimization loop. We report the implementation-independent computational scaling rather than machine-specific wall-clock numbers.

\subsection{Loss history}

\begin{figure}[htbp]
\centering
\includegraphics[width=0.65\linewidth]{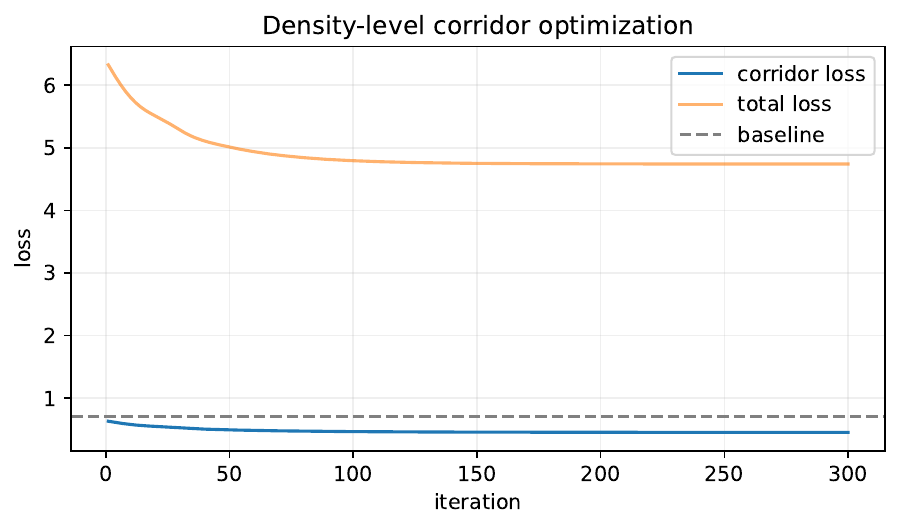}
\caption{\textbf{E1 loss history.} Corridor-alignment loss
$\mathcal L_{\mathrm{corr}}$ (blue) and total loss
$\mathcal L = 10\,\mathcal L_{\mathrm{corr}} + 0.10\,\mathcal L_\rho
            + 0.05\,\mathcal L_\beta$ (orange) versus iteration
number; baseline $\mathcal L_{\mathrm{corr}}^{\mathrm{base}} = 0.7025$
shown as the dashed grey line. The objective drops monotonically and
plateaus at $\mathcal L_{\mathrm{corr}} \approx 0.454$ by iteration
$\approx 150$, with the remaining $150$ iterations producing only
small additional refinement. Best-of-run iteration is selected by
total loss; in this run, iteration~$300$.}
\label{fig:E1_loss}
\end{figure}
Fig.~\ref{fig:E1_loss} shows the loss history. The corridor-alignment
loss decreases monotonically from $0.7025$ to $0.4542$ over the 300
iterations, a relative reduction of $35.3\%$; the total loss tracks
similarly. The plateau by iteration~$\approx 150$ indicates that the
20-parameter protocol space is essentially exhausted by the
density-level objective; we hold the optimization at $300$ iterations
to ensure convergence margin and to confirm that further iterations
do not produce drift.

\subsection{Optimized protocol parameters}

\begin{figure}[htbp]
\centering
\includegraphics[width=0.98\linewidth]{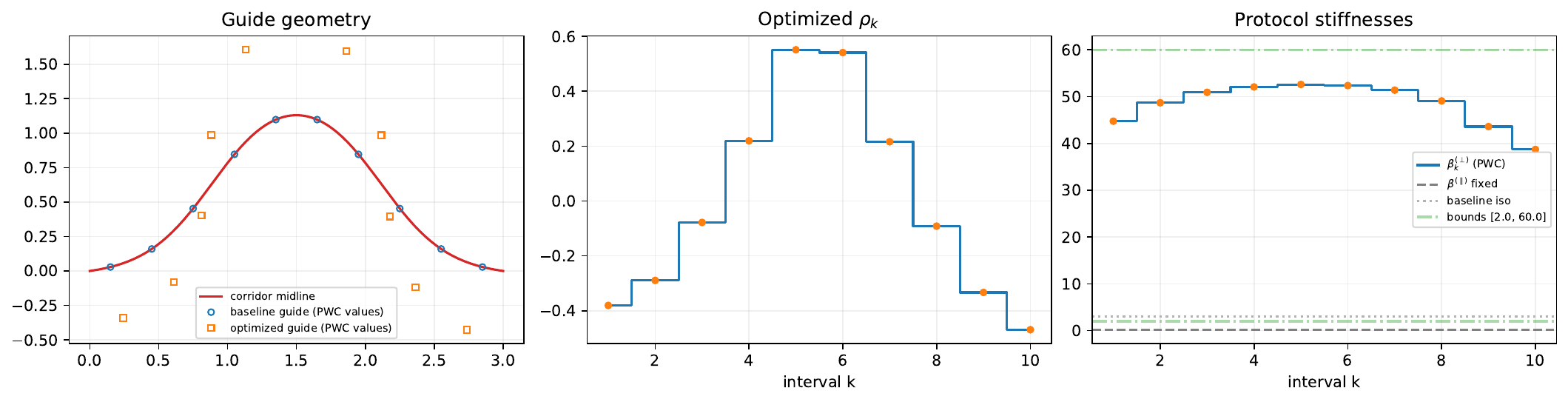}
\caption{\begin{revision}\textbf{Optimized E1 protocol parameters.}
\emph{Left:} solid red curve, corridor midline; open markers, the baseline and optimized PWC guide values. No line is drawn through the PWC values. The optimized markers lie beyond the midline around the bends to compensate for temporal lag of the cloud center.
\emph{Center:} optimized transverse offsets $\rho_k$ versus interval index $k$, shown as a step function with markers to emphasize the PWC protocol.
\emph{Right:} optimized transverse stiffness $\beta_k^{(\perp)}$ versus $k$, again shown as a PWC step function, together with the fixed longitudinal stiffness $\beta^{(\parallel)}=0.2$, the baseline isotropic $\beta=3$, and the bounds $[2,60]$.\end{revision}}
\label{fig:E1_params}
\end{figure}
\begin{revision}Fig.~\ref{fig:E1_params} plots the guide exactly as represented by the model: ten discrete PWC values in the geometry panel and step functions versus interval index. The optimized sequence changes gradually from one interval to the next while remaining discontinuous at the breakpoints. The learned transverse stiffness is about $50$ through most of the active corridor window and decreases near the terminal region where the bimodal target must be resolved.\end{revision}

\subsection{Terminal accuracy}

The terminal mode-position error and covariance error of the closed-form
marginal at $t = 1 - \varepsilon_{\mathrm{TD}}$ (with $\varepsilon_{\mathrm{TD}}
= 10^{-3}$) are reported in Table~\ref{tab:E1_terminal} for both
protocols. Terminal accuracy is essentially identical between baseline
and optimized --- as expected from Theorem~\ref{thm:closedform}, which
guarantees terminal exactness for any LQ-GM-PID protocol. The slight
increase in terminal covariance error under the optimized protocol
reflects the tighter transverse confinement and is bounded above by
the small fraction of the target covariance.

\begin{table}[t]
\centering
\small
\begin{tabular}{lcc}
\toprule
& Baseline & Optimized \\
\midrule
Corridor loss $\mathcal L_{\mathrm{corr}}$ & $0.7025$ & $0.4542$ \\
Terminal mean error & $0.0031$ & $0.0039$ \\
Terminal covariance error & $0.0008$ & $0.0025$ \\
Best iteration & --- & $300$ \\
\bottomrule
\end{tabular}
\caption{\textbf{E1 terminal accuracy summary.} Mean error and
covariance error are computed between the closed-form GMM marginal at
$t = 1 - \varepsilon_{\mathrm{TD}}$ and the prescribed target GMM,
using the standard component-wise Mahalanobis matching. Both
protocols match the target essentially exactly, as predicted by
Theorem~\ref{thm:closedform}; the small numerical residuals come from
the boundary cutoff $\varepsilon_{\mathrm{TD}}$ and from the matching
heuristic.}
\label{tab:E1_terminal}
\end{table}

\section{\textcolor{blue}{Full E2 study: finite empirical source via sample-conditioned bridges}}
\label{app:caseC}

This appendix gives the complete specification of the E2 demonstration of \S\ref{sec:E2}: source/target/corridor geometry, the coordinate-shift implementation, the optimization loss structure, optimizer configuration, the loss-history curve and the optimized protocol parameters. All quantities are reproducible from a fixed seed via \path{experiments/exp_E2_multi_entrance.ipynb}.

\subsection{Geometry, target, and source}

\paragraph{Corridor and target (identical to E1).} \begin{revision}The curved, detrended single-arch corridor $m(s)$ is exactly the one defined in App.~\ref{app:caseB}. The target is
\[
  p^{(\mathrm{tar})}(y)
  = \tfrac12\,\mathcal N\bigl(y;\,(3,+0.5),\,\mathrm{diag}(0.06,0.05)\bigr)
  + \tfrac12\,\mathcal N\bigl(y;\,(3,-0.5),\,\mathrm{diag}(0.06,0.05)\bigr).
\]
\end{revision}

\paragraph{Source distribution.} The source law Eq.~(\ref{eq:E2_source}) places two entrance modes at $(-0.3, \pm 0.5)$ with isotropic standard deviation $\sigma_0 = 0.12$. Their separation is $\Delta y=1.0$, equal to four standard deviations of either mode. The source is therefore genuinely bimodal at $t=0$.

\paragraph{Initial-particle draw.} We sample $B = 60$ initial particles $\{z^{(n)}\}_{n=1}^B$ from $p^{(\mathrm{in})}$ at the start of the experiment, with seed $42$, and hold them fixed throughout the optimization and across all diagnostic figures. This is the only random draw inside the experiment; the closed-form Riccati cascade and the density-level loss are deterministic functions of $\Gamma$ given $\{z^{(n)}\}$.

\subsection{Trainable protocol (identical to E1)}

The protocol uses $K = 10$ uniform PWC intervals on $[0,1]$, with trainable transverse offset $\rho_k \in \R$ and unconstrained $c_k \in \R$ giving the transverse stiffness via the same sigmoid reparameterization as in App.~\ref{app:caseB}. Longitudinal stiffness $\beta^{(\parallel)} = 0.2$, transverse-stiffness band $[\beta_{\min}^{(\perp)}, \beta_{\max}^{(\perp)}] = [2, 60]$, drift $\sigma_t \equiv 0$, noise schedule $\kappa_t \equiv 1$. Total parameter count: $2K = 20$.

\subsection{Coordinate-shift implementation}

\begin{revision}Because E2 has $\sigma_t=0$, the coordinate-shift shortcut of \S\ref{appA:gmm_initial} applies. The shared quadratic Green-function coefficients are computed once per protocol. For each frozen source atom $z^{(n)}$, only the linear terms and shifted target means change. The $K_{\mathrm{tar}}$ component weights are normalized \emph{within that conditional bridge}, and the $B$ normalized conditional marginals are then averaged with equal weight $1/B$. The resulting finite-source marginal therefore contains $B K_{\mathrm{tar}}=120$ Gaussian components and represents the empirical source measure exactly, while approximating the continuous source law by the fixed $B=60$ draw.\end{revision}

\paragraph{Cost decomposition.} Per gradient step there is one shared $O(Kd^3)$ Riccati cascade, a finite set of per-source linear-algebra corrections, analytic Gaussian-kernel expectations, and reverse-mode differentiation. We report the implementation-independent operation scaling rather than machine-specific wall-clock timings.

\subsection{Loss}

The total loss has the same structure as in E1:
\[
  \mathcal L(\Gamma)
  \;=\;
  \lambda_{\mathrm{corr}}\,\mathcal L_{\mathrm{corr}}^{\mathrm{E2}}
  + \lambda_\rho\,\mathcal L_\rho
  + \lambda_\beta\,\mathcal L_\beta,
\]
with weights $(\lambda_{\mathrm{corr}}, \lambda_\rho, \lambda_\beta) = (10,\,0.10,\,0.05)$, corridor-kernel widths $(\omega_\parallel, \omega_\perp) = (0.8, 0.2)$, active window $\mathcal K = \{k : s_k \le 0.80\}$, and the same $\rho$- and $c$-regularizers as in App.~\ref{app:caseB}. The corridor term Eq.~(\ref{eq:E2_loss}) averages the per-particle alignment over the $B = 60$ frozen samples; this is the only structural change from E1.

\subsection{Optimizer}

Plain gradient descent uses learning rate $\eta=3\times10^{-2}$ for $300$ iterations with no momentum. No trajectory simulation is used inside the optimization loop.

\subsection{Loss history}

\begin{figure}[htbp]
\centering
\includegraphics[width=0.65\linewidth]{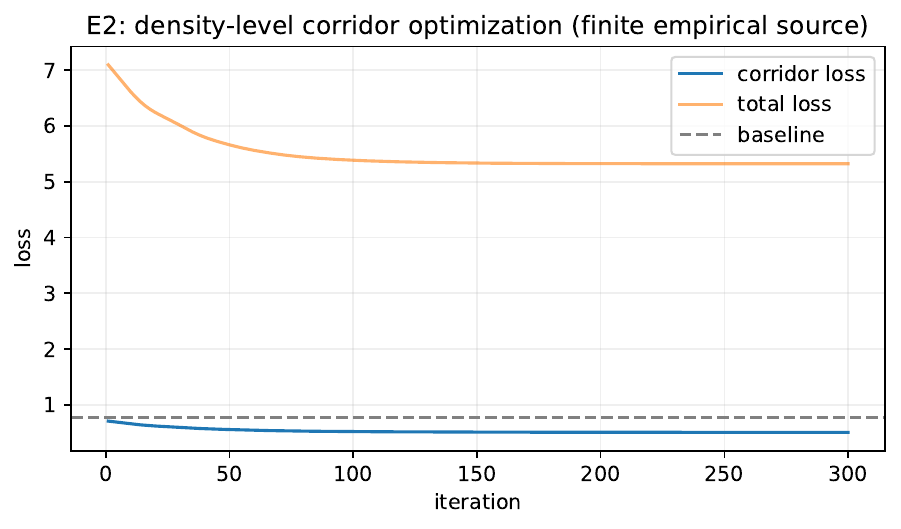}
\caption{\textbf{E2 loss history.} Corridor-alignment loss
$\mathcal L_{\mathrm{corr}}^{\mathrm{E2}}$ (blue) and total loss
(orange) versus iteration number; baseline $\mathcal L_{\mathrm{corr}}^{\mathrm{base}}
= 0.7733$ shown as the dashed grey line. As in E1, the loss decreases
monotonically and plateaus by iteration~$\approx 150$, with the
remaining 150 iterations producing only small additional refinement.
The qualitative shape mirrors Fig.~\ref{fig:E1_loss} of
App.~\ref{app:caseB}; the absolute level is slightly higher because
the finite empirical source retains the transverse spread of the sampled entrance distribution, which the corridor
kernel penalises.}
\label{fig:E2_loss}
\end{figure}
Fig.~\ref{fig:E2_loss} shows the loss history. The corridor-alignment
loss decreases monotonically from $0.7733$ to $0.5074$ over $300$
iterations, a relative reduction of $34.4\%$. The plateau structure
matches E1 closely: the 20-parameter protocol space is exhausted by
the density-level objective by iteration $\approx 150$.

\subsection{Optimized protocol parameters}

\begin{figure}[htbp]
\centering
\includegraphics[width=0.98\linewidth]{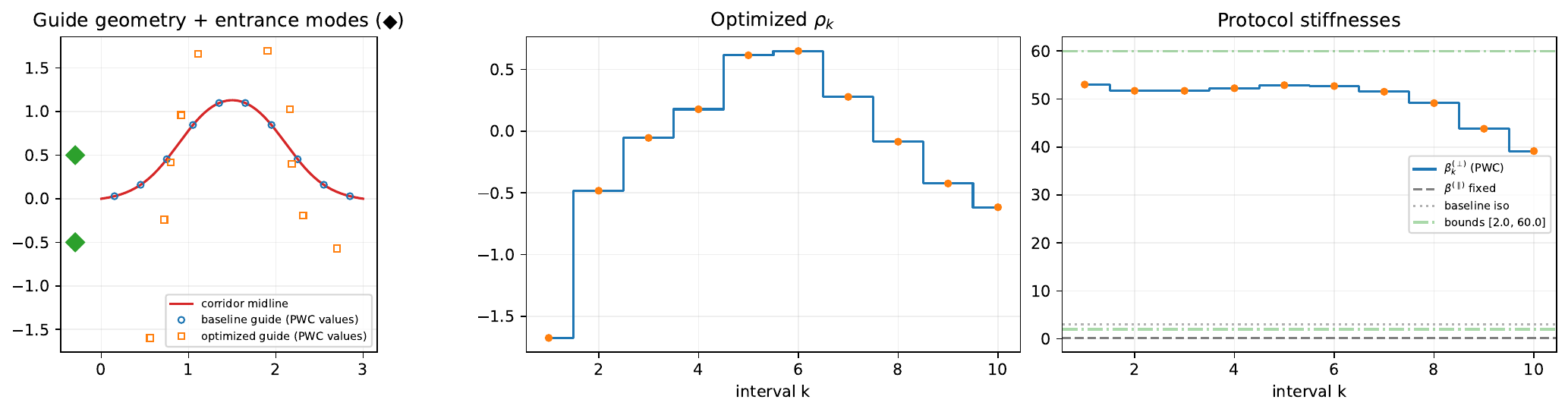}
\caption{\textbf{Optimized E2 protocol parameters.}
\emph{Left:} solid corridor midline, entrance modes (diamonds), baseline guide values (open circles), and optimized PWC guide values (open square markers); no line is drawn through either guide sequence. The optimized guide deviates more
sharply from the midline than in E1 (peak $|\rho_k| \approx 1.7$ at
the boundary intervals), reflecting the additional transport demand
from the off-axis entrance modes.
\emph{Center:} optimized $\rho_k$ vs.\ interval index $k$, shown as a PWC step function with markers. The profile changes gradually across the interval index and is approximately antisymmetric around $k=5$, but with a larger amplitude than in E1 (Fig.~\ref{fig:E1_params}, center).
\emph{Right:} optimized $\beta_k^{(\perp)}$ vs.\ $k$, shown as a PWC step function with markers, together with fixed $\beta^{(\parallel)} = 0.2$, baseline isotropic $\beta = 3$, and sigmoid bounds. The transverse stiffness profile is essentially
identical to E1: $\approx 50$ throughout the active corridor window,
dropping toward the endpoints.}
\label{fig:E2_params}
\end{figure}
Fig.~\ref{fig:E2_params} shows the optimized protocol parameters. The
key difference from E1 is in the $\rho_k$ profile (center panel):
where E1 produced a slowly varying PWC overshoot of magnitude $\approx 0.55$, E2
produces a larger overshoot ($|\rho_1| \approx 1.7$ at the boundary)
because the optimized guide must funnel particles from the off-axis
entrance positions $(-0.3, \pm 0.5)$ into the corridor by the start
of the active window. The transverse-stiffness profile (right panel)
is essentially identical to E1, indicating that the corridor
confinement requirement is determined by the corridor geometry and
target structure, not by the source distribution.

\subsection{Terminal accuracy}

\begin{table}[h!]
\centering
\small
\begin{tabular}{lcc}
\toprule
& Baseline & Optimized \\
\midrule
Corridor loss $\mathcal L_{\mathrm{corr}}^{\mathrm{E2}}$ & $0.7733$ & $0.5074$ \\
Terminal mean error & $0.0033$ & $0.0041$ \\
Terminal covariance error & $0.0008$ & $0.0025$ \\
$B$ & $60$ & $60$ \\
$B \cdot K_{\mathrm{tar}}$ (marginal mixture size) & $120$ & $120$ \\
Best iteration & --- & $300$ \\
\bottomrule
\end{tabular}
\caption{\textbf{E2 terminal accuracy summary.} Mean error and
covariance error are computed between the closed-form
$120$-component marginal at $t = 1 - \varepsilon_{\mathrm{TD}}$ and
the prescribed target GMM. Both protocols match the target to within
$5 \times 10^{-3}$ in mean error and $3 \times 10^{-3}$ in covariance
error, as predicted by Theorem~\ref{thm:closedform}; the residuals
match those of E1 (Table~\ref{tab:E1_terminal}), consistent with the conditional deterministic-start construction. These residuals do not constitute an exactness claim for the continuous non-degenerate source.}
\label{tab:E2_terminal}
\end{table}

Results of the terminal accuracy validation are reported in Table \ref{tab:E2_terminal}.

\section{Full H1 high-dimensional scaling study}
\label{app:H1}

This appendix gives the unabridged H1 study referenced from
\S\ref{sec:H1}: the mode-scaling sweep H1-B
(Fig.~\ref{fig:H1B_mode_scaling}), the closed-form vs.\ EM
subspace-variance decomposition H1-A
(Fig.~\ref{fig:H1A_subspace}), the qualitative trunk-plane and principal-component-analysis (PCA)
snapshots H1-C
(Figs.~\ref{fig:H1C_trunk}--\ref{fig:H1C_pca}), and the full
24-scenario diagnostic table (Table~\ref{tab:H1_summary}). All
quantities are reproducible from a fixed seed via
\path{experiments/exp_H1_highdim_scaling.ipynb}.

\subsection{H1-A subspace decomposition over time}

\begin{figure}[htbp]
\centering
\includegraphics[width=0.92\linewidth,height=0.55\textheight,keepaspectratio]{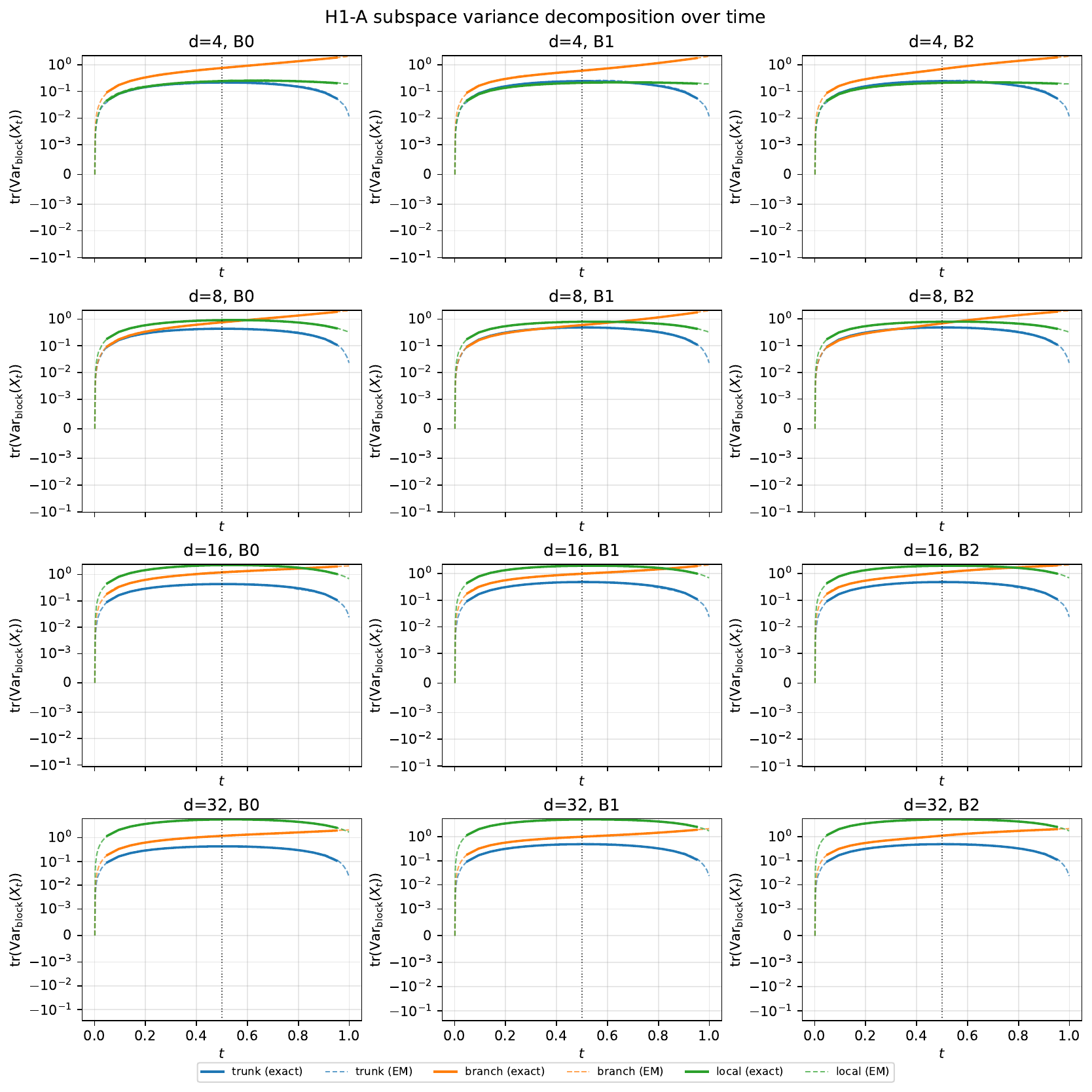}
\caption{\textbf{H1-A subspace variance decomposition over time.}
Trace of the block covariance $\mathrm{tr}(\mathrm{Var}_{\mathrm{block}}(X_t))$
for the trunk (blue), branch (orange), and local (green) blocks of
the controlled diffusion, evaluated at $21$ closed-form times (solid
lines, from \code{exact\_marginal\_gmm}) and overlaid with the
empirical EM estimate at the simulation grid (dashed,
semi-transparent). Rows: ambient dimension $d \in \{4, 8, 16, 32\}$;
columns: protocol $B_0, B_1, B_2$. Vertical dotted line marks the
design release time $t_* = 0.5$.
The exact and empirical curves agree to the resolution of the plot
in all twelve panels, validating both the closed-form marginal of
Theorem~\ref{thm:closedform} and the EM integrator. The terminal
ordering trunk $<$ branch $<$ local holds in every panel, exposing
the prescribed coarse-to-fine hierarchy. For B2 (rightmost column)
at $d \ge 16$, the branch curve visibly lags behind B0/B1 over $t \in
(0, t_*)$ --- the design-prescribed branch suppression --- and
accelerates after $t_*$, producing the ``coarse-to-fine release''
the schedule was meant to enforce.}
\label{fig:H1A_subspace}
\end{figure}

Fig.~\ref{fig:H1A_subspace} shows the subspace decomposition that
underlies the answer to Q4 in the main text. Two points are worth
noting beyond the main-text discussion. First, the agreement between
the closed-form variance trace (solid) and the EM estimate (dashed)
is a non-trivial consistency check on both the analytic apparatus of
\S\ref{appA:gmm_target_explicit} of App.~\ref{app:derivations} and
the time-stepping of the EM integrator: any error in either would
manifest as a visible separation between the two curves, and none is
present at any $(d, \mathrm{protocol})$ tested. Second, the B2
columns at $d \ge 16$ show the design-prescribed branch suppression
explicitly in the orange curves: the branch variance is held nearly
flat over $t \in (0, t_*)$ and accelerates after $t_*$, exactly
matching the protocol's piecewise-in-time stiffness profile.

\subsection{H1-B mode scaling at fixed dimension}

\begin{figure}[htbp]
\centering
\includegraphics[width=0.99\linewidth]{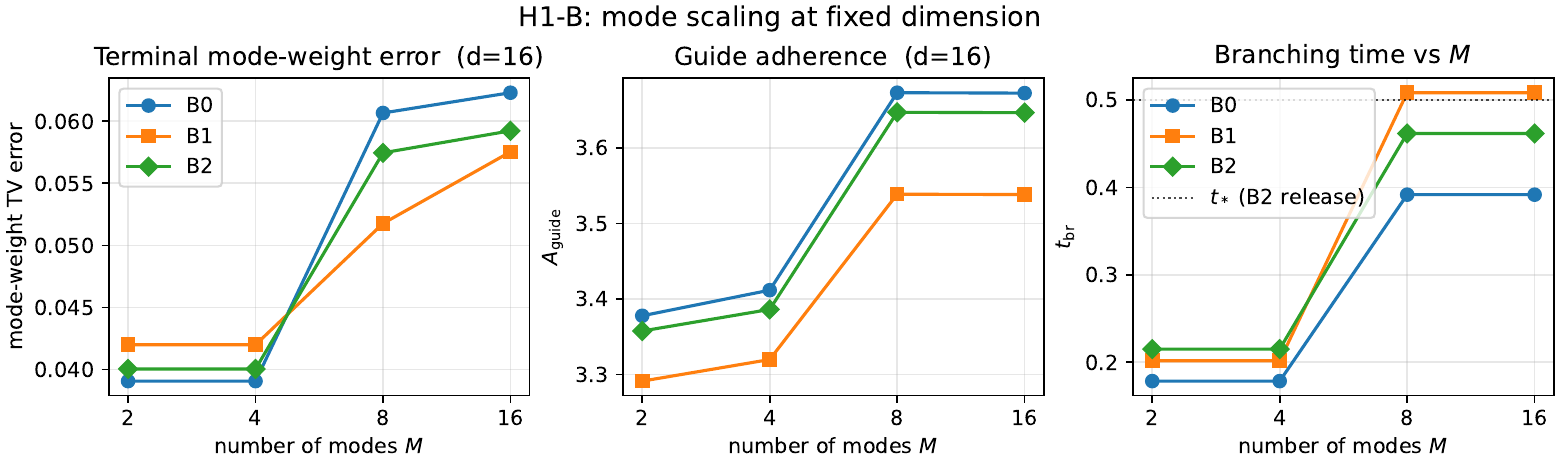}
\caption{\begin{revision}\textbf{H1-B: mode scaling at fixed $d = 16$.}
The same three diagnostics as Fig.~\ref{fig:H1_main}, now plotted versus $M\in\{2,4,8,16\}$ with $(B,L)=(2,1),(2,2),(4,2),(4,4)$.
\emph{Left:} terminal mode-weight total-variation error.
\emph{Center:} guide-adherence cost $A_{\mathrm{guide}}$; changing only the local refinement count $L$ has little effect, whereas increasing the number of coarse branches $B$ changes the cost.
\emph{Right:} branching time. The sharp change occurs when $B$ increases from $2$ to $4$, while doubling $L$ at fixed $B$ leaves $t_{\mathrm{br}}$ essentially unchanged. For $B=4$, B2 releases near the design time $t_*=0.5$.\end{revision}}
\label{fig:H1B_mode_scaling}
\end{figure}

\begin{revision}Fig.~\ref{fig:H1B_mode_scaling} addresses Q4 directly. The right panel shows that the branching-time statistic changes sharply when the number of coarse branches increases from $B=2$ to $B=4$, but is essentially unchanged when only the local refinement count $L$ is doubled. The center panel shows the analogous near-invariance of $A_{\mathrm{guide}}$ under $L$ at fixed $B$. This is the operational content of the coarse-to-fine separation claim.\end{revision}

\subsection{H1-C representative qualitative views}

\begin{figure}[htbp]
\centering
\includegraphics[width=0.99\linewidth]{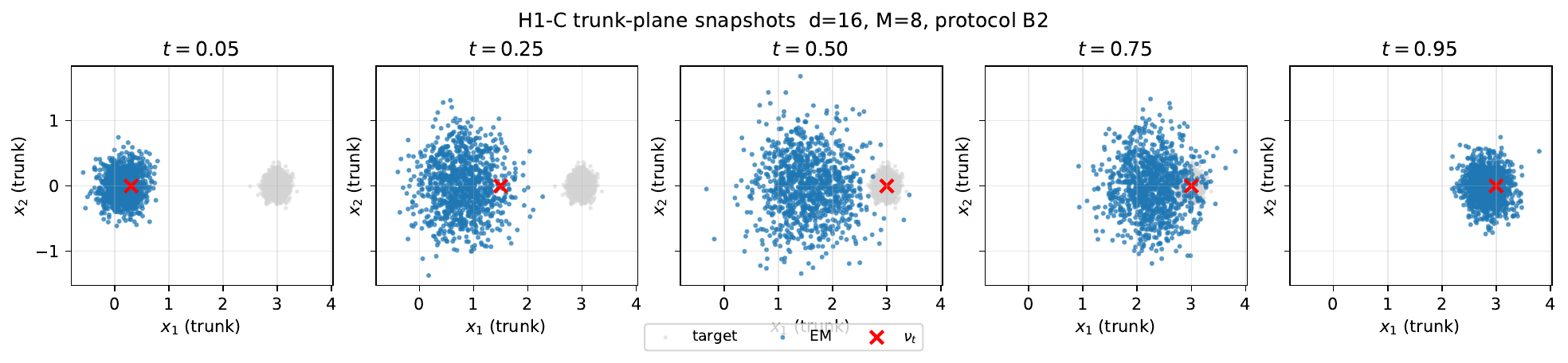}
\caption{\textbf{H1-C trunk-plane snapshots at $(d, M) = (16, 8)$
under protocol B2.} Five time slices $t \in \{0.05, 0.25, 0.50, 0.75,
0.95\}$ of $1024$ EM-sampled particles (blue) overlaid on $2000$
target samples (light grey). Trunk-plane projection $(x_1, x_2)$.
The trunk block is $2$-dimensional, and all eight target modes share
the same trunk endpoint and the same $x_2$ value, so they project
onto a single Gaussian-shaped cluster (light grey) in this plane.
The cloud (blue) starts concentrated at the origin, slides along the
trunk during $t \in [0, t_*]$ tracking the moving guide $\nu_t$ (red
cross), reaches the trunk endpoint by $t \approx t_*$, and
concentrates onto the target mode-cluster by $t = 0.95$. This view
exposes only the trunk-direction transport.}
\label{fig:H1C_trunk}
\end{figure}

\begin{figure}[htbp]
\centering
\includegraphics[width=0.99\linewidth]{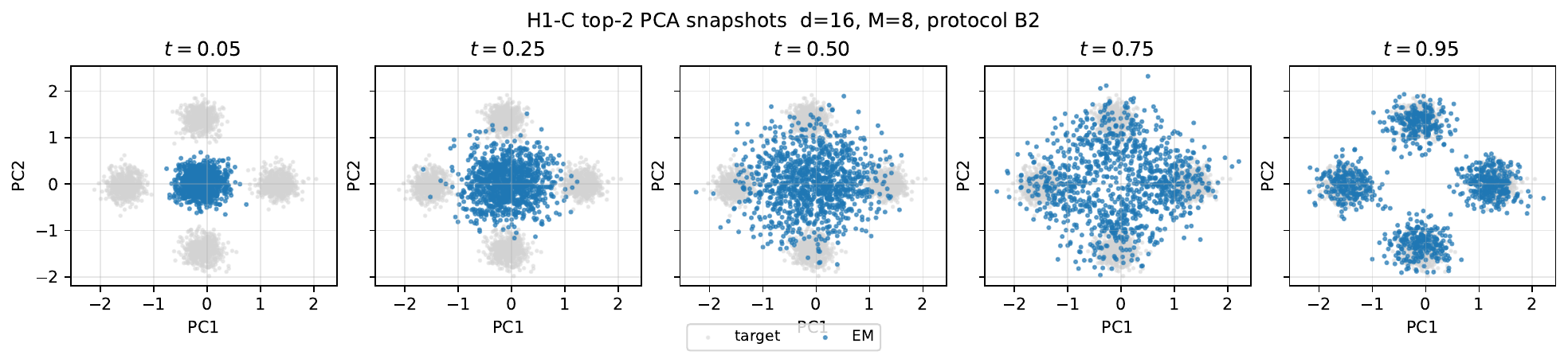}
\caption{\textbf{H1-C top-2 PCA snapshots at $(d, M) = (16, 8)$ under
protocol B2.} Same five time slices as Fig.~\ref{fig:H1C_trunk},
projected onto the top two principal components of the empirical
terminal ensemble. The PCs recover the branch plane (the two branch
coordinates carry the dominant terminal variance), so the eight
target modes appear as four corner clusters of cardinality two
(light grey), arranged in a square pattern. The EM cloud (blue)
spreads from the origin along PC2 first, then opens along PC1 after
$t \approx t_* = 0.5$, and resolves into the four corner clusters by
$t = 0.95$. This view directly visualizes the \emph{branching}
dynamics that the trunk-plane view of Fig.~\ref{fig:H1C_trunk}
obscures.}
\label{fig:H1C_pca}
\end{figure}

The two projections in Figs.~\ref{fig:H1C_trunk}--\ref{fig:H1C_pca}
are intentionally complementary: the trunk-plane view makes the
trunk-direction transport visible (and shows that the projection
collapses to a single Gaussian-shaped cluster by terminal time, as
it must, since the eight target modes share trunk coordinates), while
the PCA view isolates the branching geometry. Read together, they
confirm that LQ-GM-PID under B2 (i)~uses the first half of the time
horizon to slide along the trunk while keeping branch directions
suppressed, and (ii)~uses the second half to open the branch
directions and resolve the eight target modes. The same qualitative
two-stage behavior is implicit in the subspace traces of
Fig.~\ref{fig:H1A_subspace}; the PCA view makes it geometrically
explicit.

\begin{table}[t]
\centering\small
\begin{tabular}{llrrrr}
\toprule
$d$ & $M$ & proto & TV err. & $A_{\mathrm{guide}}$ & $t_{\mathrm{br}}$\\
\midrule
4&8&B0&0.019&1.74&0.66\\ 4&8&B1&0.024&1.78&0.76\\ 4&8&B2&0.021&1.87&0.66\\
8&8&B0&0.025&2.38&0.65\\ 8&8&B1&0.023&2.38&0.75\\ 8&8&B2&0.031&2.48&0.65\\
16&8&B0&0.061&3.67&0.39\\ 16&8&B1&0.052&3.54&0.51\\ 16&8&B2&0.058&3.65&0.46\\
32&8&B0&0.046&6.31&0.42\\ 32&8&B1&0.045&5.93&0.60\\ 32&8&B2&0.046&6.03&0.50\\
\bottomrule
\end{tabular}
\caption{\textbf{H1-A numerical summary without machine-dependent timing.} Terminal mode-weight total-variation error, guide-adherence cost, and empirical branching time for the dimension sweep at $M=8$.}
\label{tab:H1_summary}
\end{table}

\subsection{Limitations and outlook}

H1 deliberately avoids inner protocol optimization: the three protocols are hand-designed and the trunk guide $\nu_t$ is fixed. The results show that even without optimization the matrix-valued $\beta_t$ already encodes a useful high-dimensional inductive bias. Joint optimization of $\beta_t$ across blocks --- e.g.\ learning the release time $t_*$, the branch--local stiffness ratio, or making $\nu_t$ a learnable spline as in E2 --- is left to future work. We also restrict attention to $\sigma_t \equiv 0$ here; the matrix-$\sigma$ extension carries through unchanged at the algorithmic level (the forward-Riccati cascade of \S\ref{appA:pwc_closed_form} accommodates arbitrary $\sigma_k$) but the corresponding optimization story lies outside the present scaling demonstration.

\end{document}